\documentclass[10pt,twocolumn,letterpaper]{article}

\usepackage[pagenumbers]{cvpr} % To force page numbers, e.g. for an arXiv version

\usepackage{graphicx}
\usepackage{amsmath}
\usepackage{amssymb}
\usepackage{booktabs}

\makeatletter
\@namedef{ver@everyshi.sty}{}
\makeatother
\usepackage{pgfplots}
\usepackage{siunitx}
\usepackage{multirow}
\usepgfplotslibrary{colormaps} 
\pgfplotsset{compat=1.9}
\pgfplotsset{
colormap={bright}{rgb255=(0,0,0) rgb255=(78,3,100) rgb255=(2,74,255)
    rgb255=(255,21,181) rgb255=(255,113,26) rgb255=(147,213,114) rgb255=(230,255,0)
    rgb255=(255,255,255)}
}

\DeclareMathOperator*{\argmin}{arg\,min}

\usepackage[pagebackref,breaklinks,colorlinks]{hyperref}

\usepackage[capitalize]{cleveref}
\crefname{section}{Sec.}{Secs.}
\Crefname{section}{Section}{Sections}
\Crefname{table}{Table}{Tables}
\crefname{table}{Tab.}{Tabs.}

\def\cvprPaperID{*****} % *** Enter the CVPR Paper ID here
\def\confName{CVPR}
\def\confYear{2022}

\begin{document}

%%%%%%%%% TITLE - PLEASE UPDATE
\title{Poisons that are learned faster are more effective}

\author{Pedro~Sandoval-Segura$^{1}$ \quad Vasu~Singla$^{1}$ \quad Liam~Fowl$^{1}$ \quad Jonas~Geiping$^{1}$ \quad Micah~Goldblum$^{2}$\\ David~Jacobs$^{1}$ \quad Tom~Goldstein$^{1}$\\
$^{1}$University of Maryland \quad $^{2}$New York University\\
{\tt\small \{psando, vsingla, lfowl, jgeiping, dwj, tomg\}@umd.edu \quad \tt\small goldblum@nyu.edu}
% For a paper whose authors are all at the same institution,
% omit the following lines up until the closing ``}''.
% Additional authors and addresses can be added with ``\and'',
% just like the second author.
% To save space, use either the email address or home page, not both
% \and
% Second Author\\
% Second Institution \\
% {\tt\small secondauthor@i2.org}
}
\maketitle

%%%%%%%%% ABSTRACT
\begin{abstract}
   Imperceptible poisoning attacks on entire datasets have recently been touted as methods for protecting data privacy. However, among a number of defenses preventing the practical use of these techniques, early-stopping stands out as a simple, yet effective defense. To gauge poisons' vulnerability to early-stopping, we benchmark error-minimizing, error-maximizing, and synthetic poisons in terms of peak test accuracy over 100 epochs and make a number of surprising observations. First, we find that poisons that reach a low training loss faster have lower peak test accuracy. Second, we find that a current state-of-the-art error-maximizing poison is $7\times$ less effective when poison training is stopped at epoch $8$. Third, we find that stronger, more transferable adversarial attacks do not make stronger poisons. We advocate for evaluating poisons in terms of peak test accuracy.
   % the word "defense" here seems out of place since the poisoning is supposed to prevent malicious actors from scraping data
\end{abstract}

%%%%%%%%% BODY TEXT
\section{Introduction}
\label{sec:intro}

The threat of maliciously perturbed data being unexpectedly included in a dataset is high due to  automated web scraping. Web scraping is increasingly used to construct large datasets, necessary for training groundbreaking deep learning models \cite{brown2020language, radford2021learning}. The modified data, called a \textit{poison}, can induce malicious behavior in a deep neural network trained on this data by an unwitting practitioner \cite{goldblum_dataset_2021}. 
%This malicious behavior can include mis-classification of select images at test time \cite{geiping_witches_2021}, general performance degradation \cite{fowl_adversarial_2021}, or backdoor vulnerability \cite{souri2021sleeper}.
Data poisoning attacks are tailored for different kinds of erroneous model behavior: backdoor attacks ensure pre-specified input features cause inaccurate output \cite{chen2017targeted, gu2017badnets}, feature-collision attacks cause a particular target example to be misclassified as a base class \cite{shafahi_poison_2018, zhu2019transferable}, and availability attacks aim to degrade overall test performance \cite{barreno2010security, biggio2012poisoning, huang_unlearnable_2021, fowl_adversarial_2021}. In this work, we focus on \textit{availability attacks} - a flavor of data poisoning attack where the poisoner tries to induce poor performance for the victim network on the clean distribution. Often, for availability attacks, the poisoner is allowed to perturb the entire dataset, or a large portion of it. Interestingly, this form of data poisoning has recently been pitched as a means of protecting data privacy. For example, a number of practical experiments have demonstrated that adversarial poisons can be used to reduce the accuracy of protected classes within a facial recognition dataset \cite{fowl_adversarial_2021, huang_unlearnable_2021, fowl2022protecting}.

% To release a dataset securely we must prevent unwanted parties from training a model which generalizes to unseen data. For example, a social networking company might be more open to releasing a facial recognition dataset if they were confident someone else could not use the data to train their own facial recognition model. 

\pgfplotsset{
        legend cell align=left,
}

\begin{figure}[t]
    \centering
    
\begin{tikzpicture}
\begin{axis}[
    xlabel={Epochs until Loss Threshold of 0.5},
    ylabel={Peak Test Accuracy},
    xmin=0, xmax=21,
    ymin=0, ymax=100,
    xtick={0,5,10,15,20},
    ytick={0,20,40,60,80,100},
    legend pos=south east,
    legend style={nodes={scale=0.8, transform shape}, xshift=-8, yshift=65},
    ymajorgrids=true,
    grid style=dashed,
]

% 
% Dummy plots for shape legend
%
\addplot[
    only marks,
    color=black!50,
    mark=asterisk,
    mark size=2.5pt
    ]
    coordinates {
    (-1, -1) % Class-wise
    };
\addplot[
    only marks,
    color=black!50,
    mark=oplus,
    mark size=2.5pt
    ]
    coordinates {
    (-1, -1) % Sample-wise
    };
\legend{Class-wise, Sample-wise}

\addplot[
    only marks,
    color=green,
    mark=oplus, 
    mark size=2.5pt
    ]
    coordinates {
    (10, 94.3) %Clean
    };

\addplot[
    only marks,
    color=orange,
    mark=asterisk,
    mark size=2.5pt
    ]
    coordinates {
    (8, 66.6) %Classwise-random
    };

\addplot[
    only marks,
    color=magenta,
    mark=asterisk,
    mark size=2.5pt
    ]
    coordinates {
    (2, 33.2) % Unlearnable class-wise
    };
    
\addplot[
    only marks,
    color=magenta,
    mark=oplus,
    mark size=2.5pt
    ]
    coordinates {
    (4, 33.8) % Unlearnable sample-wise
    };
    
\addplot[
    only marks,
    color=violet,
    mark=oplus,
    mark size=2.5pt
    ]
    coordinates {
    (8, 57.3) % 250-step PGD, ResNet-18
    (15, 87.1) % Functional C
    (20, 83.8) % Functional CS
    (16, 80.3) % Functional CSD
    (4, 45.0) % 100-step PGD, ResNet-50
    (16, 83.4) % 10-step PGD, ResNet-50
    (10, 71.2) % 10-step MIFGSM, ResNet-18
    (8, 69.1) % 100-step MIFGSM, ResNet-18
    (9, 70.2) % 250-step MIFGSM, ResNet-18
    }; 
    
\addplot[
    only marks,
    color=cyan,
    mark=asterisk,
    mark size=2.5pt
    ]
    coordinates {
    (8, 75.8) %Regions-1
    (3, 35.0) %Regions-2
    (3, 24.5) %Regions-4
    (2, 24.5) %Regions-16
    (3, 33.2) %Regions-64
    (4, 44.3) %Regions-128
    }; 

\addplot[
    only marks,
    color=teal,
    mark=asterisk,
    mark size=2.5pt
    ]
    coordinates {
    (5, 53.5) %LowFreq-2
    (2, 43.4) %LowFreq-3
    (4, 48.6) %LowFreq-4
    (6, 47.2) %LowFreq-5
    (6, 51.0) %LowFreq-6
    (8, 53.7) %LowFreq-7
    (9, 76.0) %LowFreq-8
    };
    
% this is a dummy `axis' environment only to create the second legend
\end{axis}
\begin{axis}[
        % set some axis limits and plot the coordinates outside that box
        % so they don't show up
        xmin=1,
        xmax=2,
        ymin=1,
        ymax=2,
        % of course we also don't want to show this axis
        hide axis,
        only marks,
        legend pos=south east,
        legend style={nodes={scale=0.8, transform shape}},
    ]
    % just add some dummy plots to create the legend
    \addplot[only marks, color=green, mark=oplus*, mark size=2.5pt]
    coordinates {
    (-1, -1)}; % Clean
    
    \addplot[only marks, color=orange, mark=oplus*, mark size=2.5pt]
    coordinates {
    (-1, -1) % Classwise-random
    };
    
    \addplot[only marks, color=magenta, mark=oplus*, mark size=2.5pt]
    coordinates {
    (-1, -1) % Error-minimizing
    }; 
    
    \addplot[only marks, color=violet, mark=oplus*, mark size=2.5pt]
    coordinates {
    (-1, -1) % Error-maximizing
    }; 
    
    \addplot[only marks, color=cyan, mark=oplus*, mark size=2.5pt]
    coordinates {
    (-1,-1) % Regions
    }; 

    \addplot[only marks, color=teal, mark=oplus*, mark size=2.5pt]
    coordinates {
    (-1, -1) %Low Freq
    };
    
    \legend{Clean, Random, Error-Minimizing, Error-Maximizing, Regions, Low Frequency}
\end{axis}

\end{tikzpicture}
    \caption{Poisons which are learned faster have lower peak test accuracy on CIFAR-10. For every poison, we log both the epoch after which the training loss dips below 0.5 and the peak test accuracy over 100 epochs.}
    \label{fig:scatter-peak-acc-vs-epoch-threshold}
\end{figure}
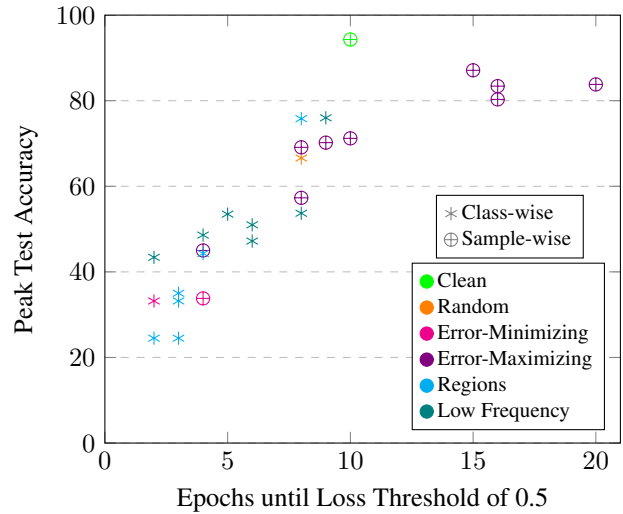

\pgfplotsset{
        legend cell align=left,
}

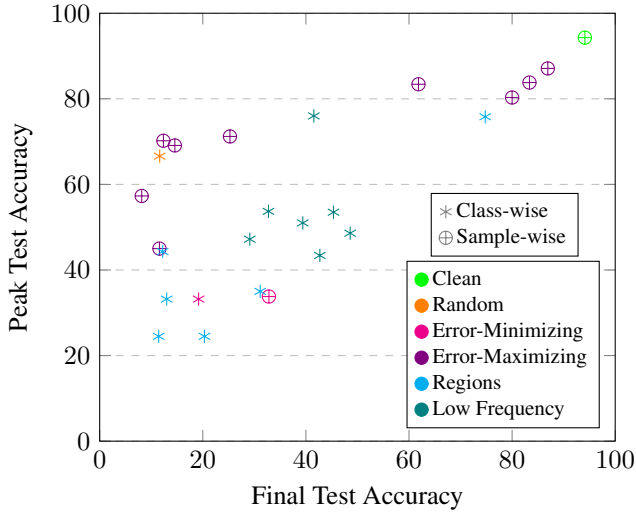
\begin{figure}[t]
    \centering
    
\begin{tikzpicture}
\begin{axis}[
    xlabel={Final Test Accuracy},
    ylabel={Peak Test Accuracy},
    xmin=0, xmax=100,
    ymin=0, ymax=100,
    xtick={0,20,40,60,80,100},
    ytick={0,20,40,60,80,100},
    legend pos=south east,
    legend style={nodes={scale=0.8, transform shape}, xshift=-8, yshift=65},
    ymajorgrids=true,
    grid style=dashed,
]

% 
% Dummy plots for shape legend
%
\addplot[
    only marks,
    color=black!50,
    mark=asterisk,
    mark size=2.5pt
    ]
    coordinates {
    (-1, -1) % Class-wise
    };
\addplot[
    only marks,
    color=black!50,
    mark=oplus,
    mark size=2.5pt
    ]
    coordinates {
    (-1, -1) % Sample-wise
    };
\legend{Class-wise, Sample-wise}

\addplot[
    only marks,
    color=green,
    mark=oplus, 
    mark size=2.5pt
    ]
    coordinates {
    (94.12, 94.3) %Clean
    };

\addplot[
    only marks,
    color=orange,
    mark=asterisk,
    mark size=2.5pt
    ]
    coordinates {
    (11.64, 66.6) %Classwise-random
    };

\addplot[
    only marks,
    color=magenta,
    mark=asterisk,
    mark size=2.5pt
    ]
    coordinates {
    (19.19, 33.2) % Unlearnable class-wise
    };
    
\addplot[
    only marks,
    color=magenta,
    mark=oplus,
    mark size=2.5pt
    ]
    coordinates {
    (32.83, 33.8) % Unlearnable sample-wise
    };
    
\addplot[
    only marks,
    color=violet,
    mark=oplus,
    mark size=2.5pt
    ]
    coordinates {
    (8.15, 57.3) % 250-step PGD, ResNet-18
    (86.93, 87.1) % Functional C
    (83.37, 83.8) % Functional CS
    (80, 80.3) % Functional CSD
    (11.59, 45.0) % 100-step PGD, ResNet-50
    (61.84, 83.4) % 10-step PGD, ResNet-50
    (25.27, 71.2) % 10-step MIFGSM, ResNet-18
    (14.58, 69.1) % 100-step MIFGSM, ResNet-18
    (12.36, 70.2) % 250-step MIFGSM, ResNet-18
    }; 
    
\addplot[
    only marks,
    color=cyan,
    mark=asterisk,
    mark size=2.5pt
    ]
    coordinates {
    (74.79, 75.8) %Regions-1
    (31.14, 35.0) %Regions-2
    (11.45, 24.5) %Regions-4
    (20.32, 24.5) %Regions-16
    (13, 33.2) %Regions-64
    (12.22, 44.3) %Regions-128
    }; 

\addplot[
    only marks,
    color=teal,
    mark=asterisk,
    mark size=2.5pt
    ]
    coordinates {
    (45.36, 53.5) %LowFreq-2
    (42.71, 43.4) %LowFreq-3
    (48.59, 48.6) %LowFreq-4
    (29.12, 47.2) %LowFreq-5
    (39.40, 51.0) %LowFreq-6
    (32.74, 53.7) %LowFreq-7
    (41.54, 76.0) %LowFreq-8
    };
    
% this is a dummy `axis' environment only to create the second legend
\end{axis}
\begin{axis}[
        % set some axis limits and plot the coordinates outside that box
        % so they don't show up
        xmin=1,
        xmax=2,
        ymin=1,
        ymax=2,
        % of course we also don't want to show this axis
        hide axis,
        only marks,
        legend pos=south east,
        legend style={nodes={scale=0.8, transform shape}},
    ]
    % just add some dummy plots to create the legend
    \addplot[only marks, color=green, mark=oplus*, mark size=2.5pt]
    coordinates {
    (-1, -1)}; % Clean
    
    \addplot[only marks, color=orange, mark=oplus*, mark size=2.5pt]
    coordinates {
    (-1, -1) % Classwise-random
    };
    
    \addplot[only marks, color=magenta, mark=oplus*, mark size=2.5pt]
    coordinates {
    (-1, -1) % Error-minimizing
    }; 
    
    \addplot[only marks, color=violet, mark=oplus*, mark size=2.5pt]
    coordinates {
    (-1, -1) % Error-maximizing
    }; 
    
    \addplot[only marks, color=cyan, mark=oplus*, mark size=2.5pt]
    coordinates {
    (-1,-1) % Regions
    }; 

    \addplot[only marks, color=teal, mark=oplus*, mark size=2.5pt]
    coordinates {
    (-1, -1) %Low Freq
    };
    
    \legend{Clean, Random, Error-Minimizing, Error-Maximizing, Regions, Low Frequency}
\end{axis}

\end{tikzpicture}
    \caption{There are a number of effective poisons for which final test accuracy doesn't correlate well with peak test accuracy. Effective poisons have both low peak test accuracy and low final test accuracy.}
    \label{fig:scatter-peak-acc-vs-final-acc}
\end{figure}

But no approach or application of data poisoning is useful if it can be circumvented easily. That is, if the victim which trains on a poison can still achieve good performance on the clean distribution, then the poison is ineffective. Many defenses have been shown to reduce a poison's effectiveness: adversarial training \cite{huang_unlearnable_2021, tao2021provable, wang_fooling_2021, geiping_what_2021}, early-stopping \cite{huang_unlearnable_2021, wallace2020concealed}, and diluting the poison with clean data \cite{huang_unlearnable_2021, fowl_adversarial_2021}. In this work, we focus on the defense of early-stopping. Using early-stopping, a practitioner chooses among the best models in terms of validation accuracy on a holdout set, over all models seen during training. In comparison to other defenses, this defense is essentially \textit{passive}. Even if the practitioner training the model is not aware of the attack (and hence would not consider actively using defenses), they would still select the best model by peak validation accuracy. %after possibly wondering about the strange overfitting behavior of their model...
The defense does not require writing new training logic, designing a new architecture, or collecting new data. 

Overall, our contributions can be summarized as follows:
\begin{itemize}
    \item We find that the early epochs of poison training are a good indicator of whether a poison will reach a high test accuracy. There exists a correlation between peak test accuracy and the number of epochs before a threshold loss is reached.
    \item We demonstrate that final test accuracy does not correlate well with peak test accuracy. We advocate for evaluating poisons based on \textit{peak test accuracy} over final test accuracy. 
    \item We find that adversarial attacks which are stronger or more transferable do not always lead to more effective error-maximizing poisons.
    % \item We craft a number of sample-wise, class-wise, error-minimizing, error-maximizing, and synthetic random poisons, and train we train on them to provide a more complete picture of poisoning.
    %classical vision networks like ResNet \cite{he2016deep}, but also newer architectures like ViT \cite{dosovitskiy_image_2021} and ConvMixer \cite{trockman2022patches},

\end{itemize}

\section{Related Work}

To conduct availability attacks on neural networks, recent works have modified data to explicitly cause gradient vanishing \cite{shen2019tensorclog} or have minimized the loss with respect to the input image \cite{huang_unlearnable_2021}. More recently, strong adversarial attacks, which perturb clean data by maximizing the loss with respect to the input image, have been shown to be the most successful approach thus far \cite{fowl_adversarial_2021}. But success has largely been defined by the final test set accuracy of the poisoned network. In this work, we show that final test set accuracy does not tell the whole story by investigating early-stopping as it is the simplest and most practical mitigation. A thorough overview of data poisoning methods can be found in \cite{goldblum_dataset_2021}.

\subsection{Error-Minimizing Noise}

Dubbed \textit{unlearnable examples}, images perturbed with error-minimizing noises are a surprisingly good data poisoning attack. A ResNet-18 network trained on a CIFAR-10 \cite{krizhevsky2009learning} sample-wise error-minimizing poison achieves $19.9\%$ final test accuracy, while the class-wise variant achieves $16.4\%$ final test accuracy after $60$ epochs of training \cite{huang_unlearnable_2021}. The discovery of unlearnable examples also included analysis of error-maximizing noise, but results demonstrated that error-maximizing noise was not as effective in degrading test set performance of a classifier.

\subsection{Error-Maximizing Noise}

A number of works have shown that adversarial examples can fool DNNs at test time \cite{szegedy_intriguing_2014, goodfellow_explaining_2015, kurakin_adversarial_2017, carlini_towards_2017, madry_towards_2019} through the use of error-maximizing perturbations. Adversarial examples are crafted by optimizing an objective which seeks to maximize the network's prediction error. While many adversarial attacks operate under an additive threat model, where the adversary is allowed to add an imperceptible perturbation vector to a clean image, spatial \cite{xiao_spatially_2018} and functional \cite{laidlaw_functional_2019} threat models have also been explored. Functional adversarial attacks introduced a novel class of threat models which, at the time, produced the highest attack success rates even after adversarial training. The ReColorAdv \cite{laidlaw_functional_2019} attack uses a parameterized function to map each color in an image to a new color in the adversarial example. Laidlaw \etal also explore combining this functional threat model with spatial and additive threat models, resulting in stronger, more transferable attacks. In \cref{subsection:stronger-attacks}, we explore whether poisons created using these attacks are effective.

The use of error-maximizing noises for data poisoning is highly effective as an availability attack. The most effective error-maximizing poison can poison a network to achieve $6.25\%$ test accuracy on CIFAR-10 \cite{fowl_adversarial_2021}. In \cref{subsection:poisons-learned-faster}, we find that if the victim network were to train a network for fewer epochs, this poison is less effective.

\begin{figure*}[ht]
     \centering
     \begin{subfigure}[b]{0.24\textwidth}
         \centering
         \begin{subfigure}{\textwidth}
            \centering
            \includegraphics[width=0.45\textwidth]{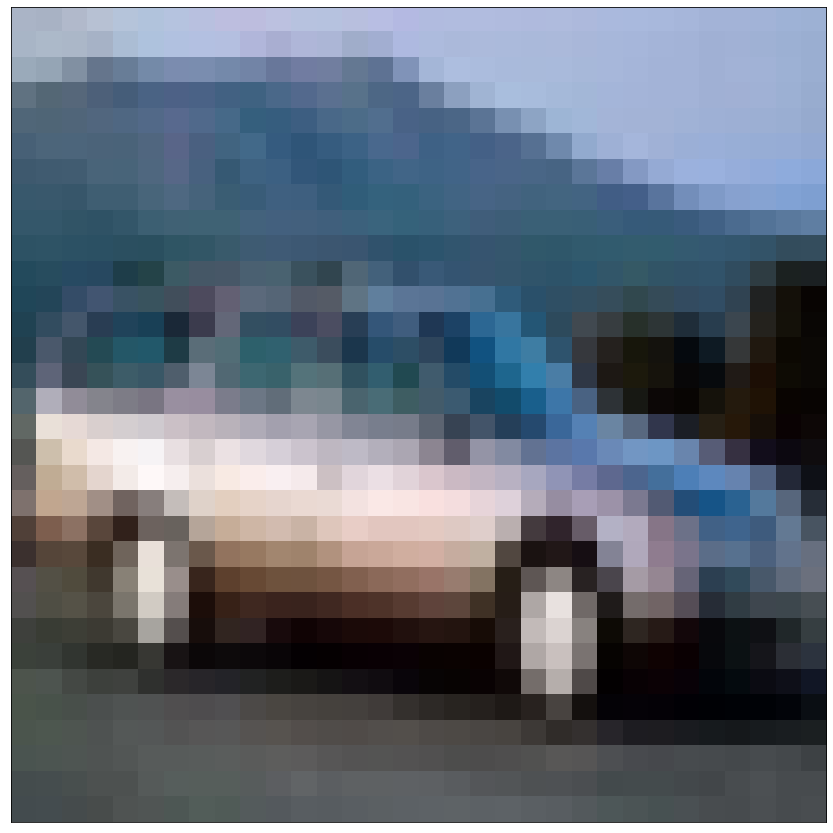}
            \caption{Clean}
            \label{fig:example-poison-clean}
          \end{subfigure}

        \includegraphics[width=\textwidth]{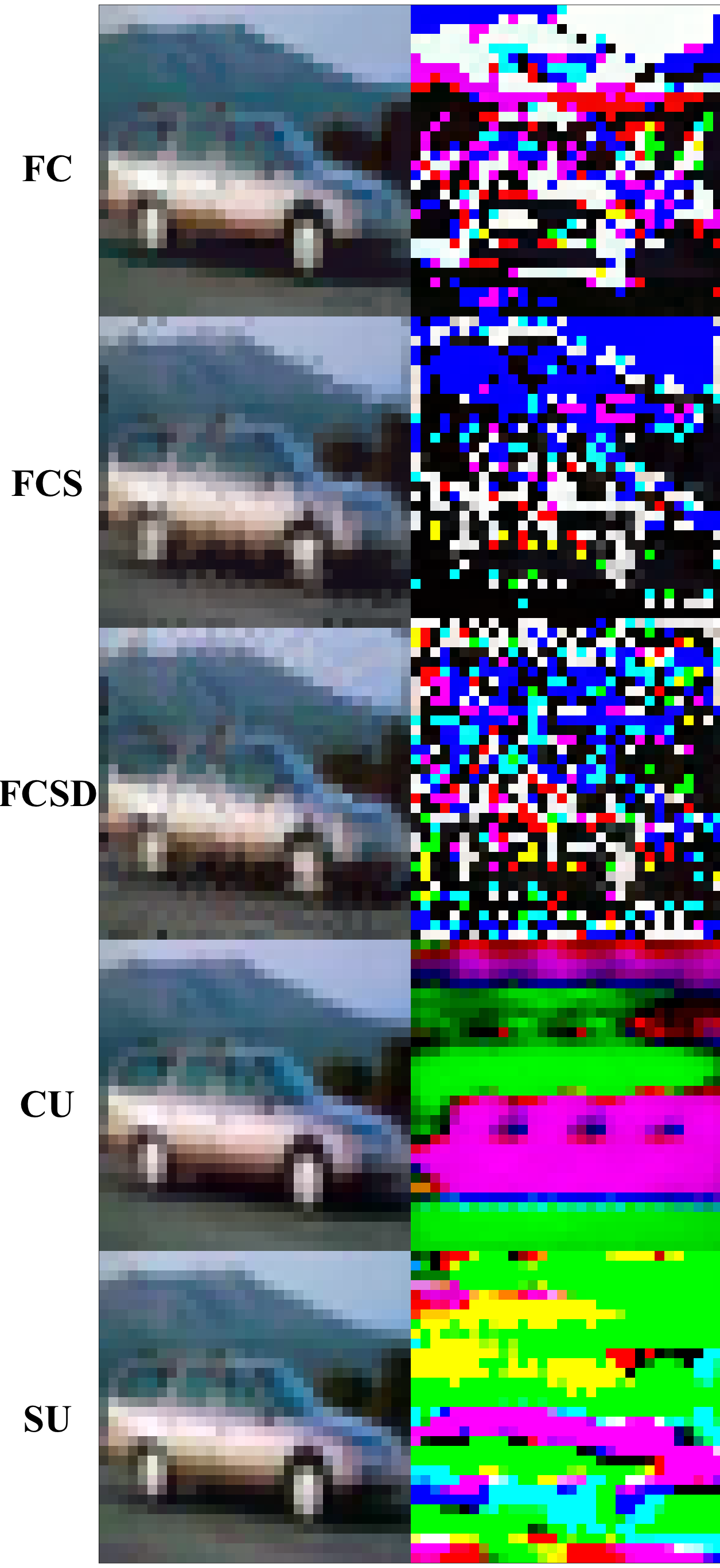}
        \caption{Adversarial}
        \label{fig:example-poison-adv-1}
     \end{subfigure}
     \hfill
     \begin{subfigure}[b]{0.24\textwidth}
         \centering
         \includegraphics[width=\textwidth]{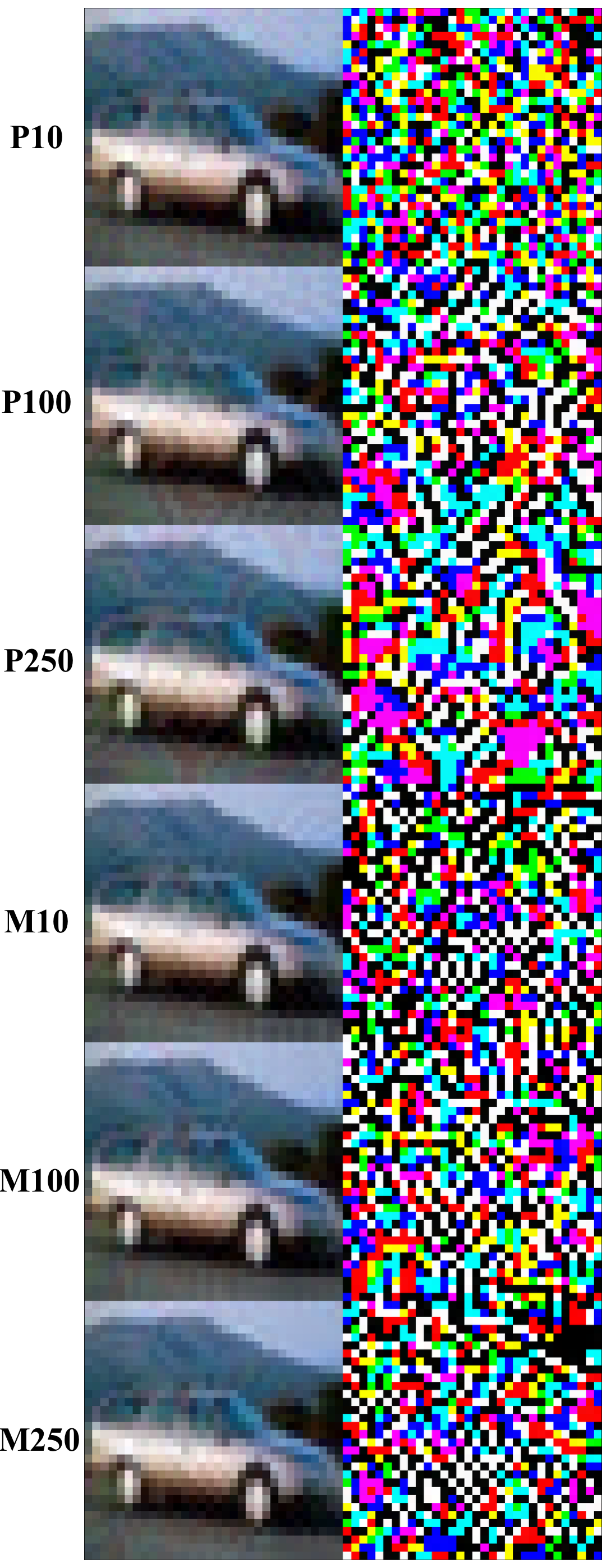}
         \caption{Adversarial}
         \label{fig:example-poisons-adv-2}
     \end{subfigure}
     \hfill
     \begin{subfigure}[b]{0.24\textwidth}
         \centering
         \includegraphics[width=\textwidth]{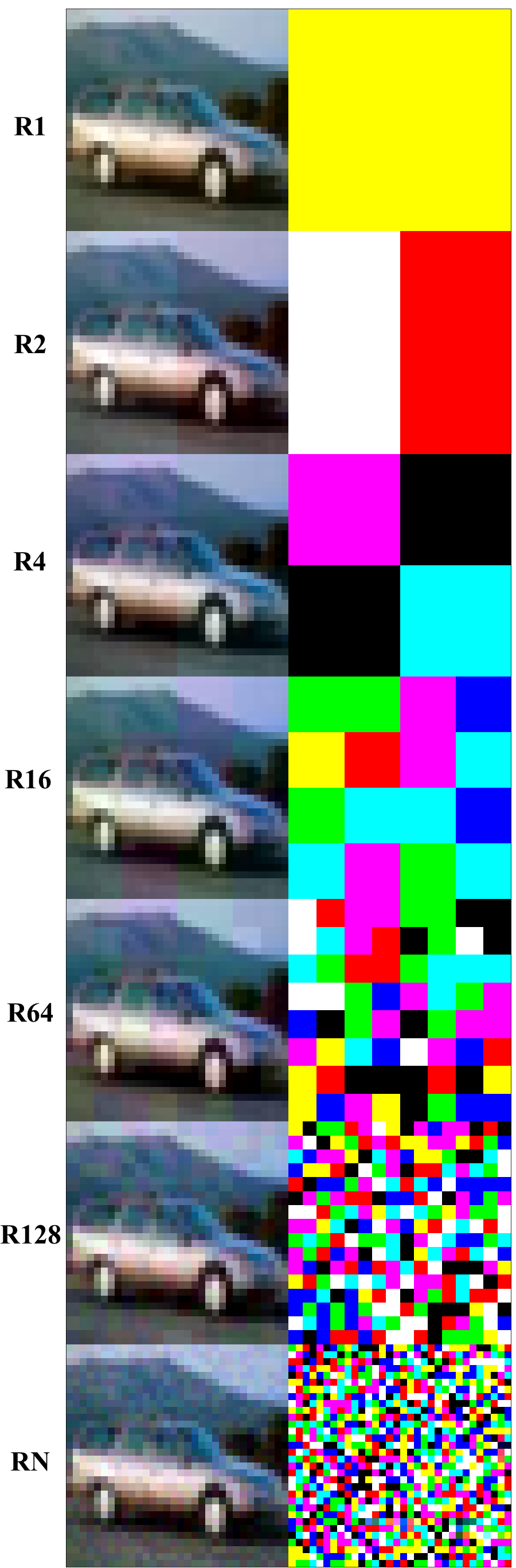}
         \caption{Regions}
         \label{fig:example-poison-regions}
     \end{subfigure}
     \hfill
     \begin{subfigure}[b]{0.24\textwidth}
         \centering
         \includegraphics[width=\textwidth]{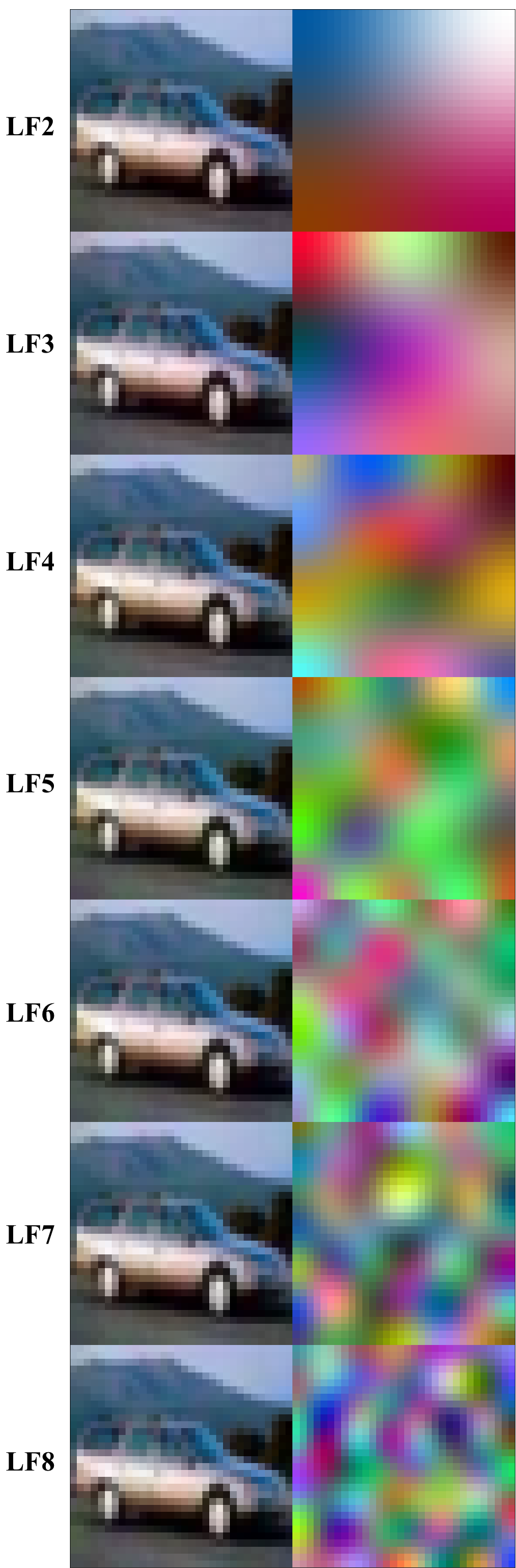}
         \caption{Low Frequency}
         \label{fig:example-poison-lowfreq}
     \end{subfigure}
     \caption{A CIFAR-10 image and its corresponding normalized perturbation from different poisons labeled by IDs from \cref{tab:summary-all-poisons}. 
     The original clean sample is shown in \protect\subref{fig:example-poison-clean}. Error-minimizing and error-maximizing poisons are shown in columns \protect\subref{fig:example-poison-adv-1} and \protect\subref{fig:example-poisons-adv-2}. Synthetic random poisons are shown in columns \protect\subref{fig:example-poison-regions} and \protect\subref{fig:example-poison-lowfreq}.
     }
    \label{fig:images-poisons}
\end{figure*}

\section{Adversarial and Synthetic Poisons}

\subsection{Problem Statement}

We formulate the problem of creating a clean-label poison in the context of image classification with DNNs, following \cite{huang_unlearnable_2021}. For a $K$-class classification task, we denote the clean training and test datasets as $\mathcal{D}_{c}$ and $\mathcal{D}_{t}$, respectively. We assume $\mathcal{D}_{c}, \mathcal{D}_{t} \sim \mathcal{D}$. We let $f_{\theta}$ represent a classification DNN with parameters $\theta$. The goal is to perturb $\mathcal{D}_{c}$ into a poison $\mathcal{D}_{p}$ such that when DNNs are trained on $\mathcal{D}_{p}$, they perform poorly on test set $\mathcal{D}_{t}$.

Suppose there are $n$ samples in the clean training set, i.e. $\mathcal{D}_{c} = \{(x_i, y_i)\}_{i=1}^{n}$ where $x_i \in \mathbb{R}^{d}$ are the inputs and $y_i \in \{1,..., K\}$ are the labels. We denote the poisoned dataset as $\mathcal{D}_{p} = \{(x_{i}', y_i)\}_{i=1}^{n}$ where $x_{i}' = x_{i} + \delta$ is the poisoned version of the example $x_{i} \in \mathcal{D}_{c}$ and where $\delta \in \Delta \subset \mathbb{R}^d$ is the  perturbation\footnote{Functional poisons are not contained within this additive threat model. Whereas an additive attack perturbs each pixel separately, functional attacks apply a function $f(\cdot)$ to every pixel, and the adversarial example is written $x_{i}' = f(x_i)$.}. Perturbations $\delta$ are bounded by $\|\delta \|_{p} < \epsilon$ where $\|\cdot\|_{p}$ is the $\ell_{p}$ norm and $\epsilon$ is set to be small enough that it does not affect the utility of the example. Perturbations are typically restricted to be sampled from a set of allowable perturbations $\Delta$ which, in effect, defines the threat model.

Poisons are created by applying a perturbation to a clean image in either a class-wise or sample-wise manner. When a perturbation is applied class-wise, every sample of a given class is perturbed in the same way. That is, $x_{i}' = x_{i} + \delta_{y_i}$ and $\delta_{y_i} \in \Delta_{C} = \{\delta_{1},..., \delta_{K} \}$. Due to the explicit correlation between the perturbation and the true label, it should not be surprising that class-wise poisons appear to trick the model to learn the perturbation over the image content, subsequently reducing generalization to the clean test set. When a poison is applied sample-wise, every sample of the training set is perturbed independently. That is, $x_{i}' = x_{i} + \delta_{i}$ and $\delta_{i} \in \Delta_{S} = \{\delta_{1},..., \delta_{n} \}$.

Error-maximizing availability poisoning aims to solve the following objective in terms of perturbations $\delta_i \in \Delta$ to samples $x_i \in \mathcal{D}_c$:

\begin{equation}
    \max_{\delta \in \Delta} \mathbb{E}_{(x,y) \sim \mathcal{D}_t} \left[ \mathcal{L}(f(x), y; \theta(\delta)) \right]
    \label{eq:max-expected-loss}
\end{equation}

\noindent
whereas error-minimizing poisons solve the following objective:

\begin{equation}
    \min_{\delta \in \Delta} \mathbb{E}_{(x,y) \sim \mathcal{D}_t} \left[ \mathcal{L}(f(x), y; \theta(\delta)) \right].
    \label{eq:min-expected-loss}
\end{equation}

\noindent
Both objectives are bi-level optimization problems, as $\theta$ is implicitly defined via:

\begin{equation}
    \theta(\delta) = \argmin_{\theta} \sum_{(x_i, y_i) \in \mathcal{D}_c} \mathcal{L}(f(x_i + \delta_i), y_i; \theta)
    \label{eq:min-empirical-train-loss}
\end{equation}

In the case of an error-maximizing poison, the adversary intends for a network, $f$, trained on the poison in the manner of \cref{eq:min-empirical-train-loss}, to perform poorly on the test distribution $D_t$, as in \cref{eq:max-expected-loss}, from which $D_c$ was also sampled. We use SGD to optimize \cref{eq:min-empirical-train-loss} and projected gradient descent (PGD) to optimize \cref{eq:max-expected-loss} or \cref{eq:min-expected-loss}, following \cite{huang_unlearnable_2021} and \cite{fowl_adversarial_2021}.

%In the case of an error-minimizing poison, the adversary intends for a network, $f$, trained on the poison in the manner of \cref{eq:min-empirical-train-loss}, to perform exceedingly well on the test distribution $D_t$, as in \cref{eq:min-expected-loss}. 

With the exception of the functional poisons, all poisons we consider contain perturbations $\delta$ which satisfy $\|\delta\|_{\infty} \leq \epsilon = \frac{8}{255}$, as previous studies have determined that these perturbations are imperceptible to human observers \cite{huang_unlearnable_2021}. All poisons are outlined in \cref{tab:summary-all-poisons}. For the remainder of our work, we refer to a poison by their ID. Only \textbf{P10} and \textbf{P100} are optimized using a ResNet-50. All other poisons are crafted using a ResNet-18 \cite{he2016deep}, trained from scratch on clean CIFAR-10 for $40$ epochs. 
We plot a sample image and its corresponding normalized perturbation for a variety of poisons in \cref{fig:images-poisons}.

% Experiments by Fowl \etal using targeted adversarial attacks suggest that while error-maximizing sample-wise perturbations are unique, target label mappings are leaked \cite{fowl_adversarial_2021}. Using a bit of extra clean data it is possible to expose the target labels used during the adversarial attack. 

\subsection{Generating Adversarial Poisons}

We craft % still sad that "brewing" hasnt been more widely adopted :>
a number of error-maximizing poisons using the open-source code of Fowl  \etal~\cite{fowl_adversarial_2021}. In particular, we employ targeted 10, 100, and 250-step PGD \cite{madry_towards_2019} and MI-FGSM \cite{dong_boosting_2018} attacks. The poison crafted using a targeted 250-step PGD attack is the main poison published by Fowl \etal~\cite{fowl_adversarial_2021}, where a ResNet-18 trained on the poison will achieve a final test set accuracy of $6.25\%$. Our version\footnote{Fowl \etal only release the untargeted version of their poison, so we crafted the targeted version using their open-source code.} of this poison is slightly less effective, achieving final $8.2\%$ test accuracy. Our functional poisons make use of a functional attack, ReColorAdv (C) \cite{laidlaw_functional_2019}, which can be combined with a spatial attack, StAdv (S) \cite{xiao_spatially_2018} and an additive delta (D) attack like PGD. Following the naming style from Laidlaw \etal, the \textbf{FCSD} poison perturbs images using ReColor, StAdv, and PGD. The functional poisons are the only poisons where perturbations are not constrained to an $\ell_{\infty}$ ball around the original image. Functional attacks use different distance measures to quantify perceptual similarity between the clean and the adversarial example.

We also craft two error-minimizing poisons, \textbf{SU} and \textbf{CU}, using open-source code from Huang \etal~\cite{huang_unlearnable_2021}, one of which is sample-wise and the other class-wise.

\begin{table}[t]
  \centering
  \resizebox{3.3in}{!}{
  \begin{tabular}{llccll}
    \toprule
    ID & Detail & Type  & Appl. & Peak               &  Final \\
       &        &       &       & Acc ($\downarrow$) &  Acc ($\downarrow$) \\
    \midrule
    P10 & 10-step PGD & Max & S & 83.43 & 61.84 \\
    P100 & 100-step PGD & Max & S & 45.02 & 11.59 \\
    P250 & 250-step PGD \cite{fowl_adversarial_2021} & Max & S & 57.30 &  \textbf{8.15} \\
    
    M10 & 10-step MIFGSM & Max & S & 71.18 & 25.27 \\
    M100 & 100-step MIFGSM & Max & S & 69.05 & 14.58 \\ 
    M250 & 250-step MIFGSM & Max & S & 70.20 & 12.36 \\
    
    FC & 100-step Functional C & Max & S & 87.09 & 86.93 \\ 
    FCS & 100-step Functional C+S & Max & S & 83.82 & 83.37 \\
    FCSD & 100-step Functional C+S+D & Max & S & 80.31 & 80.00 \\

    SU & SW-Unlearnable \cite{huang_unlearnable_2021} & Min & S & \textbf{33.77} & 32.83 \\
    CU & CW-Unlearnable \cite{huang_unlearnable_2021} & Min & C & 33.22 & 19.19 \\
    R1 & 1 region & R & C & 75.83 & 74.79 \\
    R2 & 2 regions & R & C & 34.96 & 31.14 \\
    R4 & 4 regions & R & C & \textbf{24.51} & \textbf{11.45} \\
    R16 & 16 regions & R & C & 24.54 & 20.32 \\
    R64 & 64 regions & R & C & 33.21 & 13.00 \\
    R128 & 128 regions & R & C & 44.26 & 12.22 \\
    RN & 1024 regions & R & C & 66.58 & 11.64 \\
    
    LF2 & 2x2 DCT & R & C & 53.51 & 45.36 \\
    LF3 & 3x3 DCT & R & C & 43.36 & 42.71 \\
    LF4 & 4x4 DCT & R & C & 48.59 & 48.59 \\
    LF5 & 5x5 DCT & R & C & 47.16 & 29.12 \\
    LF6 & 6x6 DCT & R & C & 51.04 & 39.40 \\
    LF7 & 7x7 DCT & R & C & 53.72 & 32.74 \\
    LF8 & 8x8 DCT & R & C & 75.96 & 41.54 \\

    \bottomrule
  \end{tabular}
  } % end resizebox
  \caption{A summary of all CIFAR-10 poisons considered in this work. Poisons are given an ID for easy reference. Poison perturbations are either Error-Minimizing (Min), Error-Maximizing (Max), or Random (R), and are applied to clean images in a Class-wise (C) or Sample-wise (S) manner. The best class-wise and sample-wise poisons are in bold.}
  \label{tab:summary-all-poisons}
\end{table}

\subsection{Generating Synthetic Poisons}

A quick glance at the normalized error-minimizing and error-maximizing perturbations in \cref{fig:images-poisons} illustrates the difficulty of understanding why they work. A much simpler noise type to understand is that of synthetic, random perturbations. Our synthetic poisons are randomly generated and are designed to take advantage of two kinds of features that convolutional networks seem to be biased towards: spatially-local \cite{baker2020local} and low frequency \cite{ortiz2020holdmetight, guo2018low}. Both these synthetic class-wise noises represent a first step in designing poisons which are learned quickly.

\textbf{Regions noise.} To generate a noise with $n$ patch regions, we sample $n$ vectors of size $3$ from a Bernoulli distribution. Each vector is then scaled to lie in the range $[-\frac{8}{255}, \frac{8}{255}]$. Finally, the $n$ vectors are repeated along two dimensions to achieve a shape of $32 \times 32 \times 3$. With the exception of the \textbf{R2} poison, a Regions noise contains patches of size $\frac{32}{\sqrt{n}} \times \frac{32}{\sqrt{n}}$. 

%Our Regions perturbations are similar to those of \cite{yu_indiscriminate_2021}, but we scale our perturbations to reach the edge of the $\epsilon$-ball. 
%Our perturbations are also not necessarily linearly separable. In fact, the $10$ perturbations of the \textbf{R1} poison are not linearly separable. 

\textbf{Low frequency noise.} To generate low frequency perturbations, we first sample an $n \times n \times 3$ matrix of Gaussian noise, representing a DCT block. We then append zeros to the matrix along the first two dimensions to achieve a shape of $32 \times 32 \times 3$ and subsequently perform the inverse DCT transform. When $n$ is small, the resulting image contains low frequency patterns. 

% Figure design:
% Loss Plot 1, Loss Plot 2, Loss Plot 3
% Test Plot 1, Test Plot 2, Test Plot 3
\begin{figure*}[t]
    \centering
    \begin{subfigure}[b]{0.28\textwidth}
        \centering
        \resizebox{\textwidth}{!}{
            \begin{tikzpicture}
\begin{axis}[
    xlabel={Epoch},
    ylabel={Train Loss},
    xmin=0, xmax=20,
    ymin=0, ymax=2.5,
    xtick={0,5,10,15,20},
    ytick={0,0.5,1,1.5,2,2.5},
    legend pos=north east,
    legend style={nodes={scale=0.7, transform shape}},
    ymajorgrids=true,
    grid style=dashed,
    every axis plot/.append style={very thick}
]

\addplot[
    color=green,
    ]
    coordinates { %Clean
(0, 1.837)
(1, 1.562)
(2, 1.359)
(3, 1.061)
(4, 0.960)
(5, 0.841)
(6, 0.788)
(7, 0.587)
(8, 0.601)
(9, 0.551)
(10, 0.446)
(11, 0.472)
(12, 0.457)
(13, 0.377)
(14, 0.348)
(15, 0.373)
(16, 0.311)
(17, 0.365)
(18, 0.299)
(19, 0.326)
(20, 0.298)
(21, 0.249)
(22, 0.236)
(23, 0.260)
(24, 0.280)
(25, 0.229)
(26, 0.179)
(27, 0.202)
(28, 0.226)
(29, 0.212)
(30, 0.218)
(31, 0.199)
(32, 0.162)
(33, 0.205)
(34, 0.158)
(35, 0.184)
(36, 0.193)
(37, 0.164)
(38, 0.187)
(39, 0.146)
(40, 0.169)
(41, 0.150)
(42, 0.148)
(43, 0.187)
(44, 0.158)
(45, 0.156)
(46, 0.154)
(47, 0.166)
(48, 0.164)
(49, 0.198)
(50, 0.049)
(51, 0.062)
(52, 0.035)
(53, 0.028)
(54, 0.023)
(55, 0.020)
(56, 0.019)
(57, 0.014)
(58, 0.016)
(59, 0.011)
(60, 0.011)
(61, 0.019)
(62, 0.012)
(63, 0.009)
(64, 0.008)
(65, 0.011)
(66, 0.012)
(67, 0.009)
(68, 0.007)
(69, 0.006)
(70, 0.007)
(71, 0.008)
(72, 0.010)
(73, 0.008)
(74, 0.006)
(75, 0.003)
(76, 0.008)
(77, 0.007)
(78, 0.005)
(79, 0.004)
(80, 0.005)
(81, 0.006)
(82, 0.003)
(83, 0.005)
(84, 0.003)
(85, 0.006)
(86, 0.003)
(87, 0.004)
(88, 0.004)
(89, 0.004)
(90, 0.004)
(91, 0.005)
(92, 0.004)
(93, 0.005)
(94, 0.002)
(95, 0.003)
(96, 0.003)
(97, 0.003)
(98, 0.003)
(99, 0.003)
    };
    
\addplot[
    color=red,
    ]
    coordinates { %Class-wise Random
(0, 1.953)
(1, 1.621)
(2, 1.428)
(3, 1.112)
(4, 1.005)
(5, 0.805)
(6, 0.807)
(7, 0.558)
(8, 0.242)
(9, 0.077)
(10, 0.026)
(11, 0.013)
(12, 0.016)
(13, 0.005)
(14, 0.003)
(15, 0.005)
(16, 0.002)
(17, 0.001)
(18, 0.002)
(19, 0.001)
(20, 0.002)
(21, 0.005)
(22, 0.003)
(23, 0.001)
(24, 0.005)
(25, 0.003)
(26, 0.004)
(27, 0.001)
(28, 0.003)
(29, 0.001)
(30, 0.001)
(31, 0.001)
(32, 0.001)
(33, 0.002)
(34, 0.019)
(35, 0.005)
(36, 0.002)
(37, 0.001)
(38, 0.001)
(39, 0.001)
(40, 0.001)
(41, 0.001)
(42, 0.003)
(43, 0.003)
(44, 0.003)
(45, 0.002)
(46, 0.001)
(47, 0.001)
(48, 0.001)
(49, 0.001)
(50, 0.001)
(51, 0.001)
(52, 0.001)
(53, 0.001)
(54, 0.001)
(55, 0.001)
(56, 0.001)
(57, 0.001)
(58, 0.001)
(59, 0.001)
(60, 0.001)
(61, 0.001)
(62, 0.001)
(63, 0.001)
(64, 0.001)
(65, 0.001)
(66, 0.001)
(67, 0.001)
(68, 0.001)
(69, 0.001)
(70, 0.001)
(71, 0.001)
(72, 0.001)
(73, 0.001)
(74, 0.001)
(75, 0.001)
(76, 0.001)
(77, 0.001)
(78, 0.001)
(79, 0.001)
(80, 0.001)
(81, 0.001)
(82, 0.001)
(83, 0.001)
(84, 0.001)
(85, 0.001)
(86, 0.001)
(87, 0.001)
(88, 0.001)
(89, 0.001)
(90, 0.001)
(91, 0.001)
(92, 0.001)
(93, 0.001)
(94, 0.001)
(95, 0.001)
(96, 0.001)
(97, 0.001)
(98, 0.001)
(99, 0.001)
    };
    
\addplot[
    color=blue,
    ]
    coordinates { % Class-wise Unlearnable
(0, 1.911)
(1, 1.357)
(2, 0.289)
(3, 0.139)
(4, 0.092)
(5, 0.056)
(6, 0.056)
(7, 0.041)
(8, 0.043)
(9, 0.038)
(10, 0.037)
(11, 0.031)
(12, 0.025)
(13, 0.017)
(14, 0.023)
(15, 0.018)
(16, 0.016)
(17, 0.027)
(18, 0.018)
(19, 0.018)
(20, 0.024)
(21, 0.033)
(22, 0.027)
(23, 0.011)
(24, 0.020)
(25, 0.014)
(26, 0.015)
(27, 0.030)
(28, 0.021)
(29, 0.010)
(30, 0.016)
(31, 0.015)
(32, 0.026)
(33, 0.012)
(34, 0.014)
(35, 0.016)
(36, 0.016)
(37, 0.008)
(38, 0.017)
(39, 0.010)
(40, 0.008)
(41, 0.009)
(42, 0.014)
(43, 0.016)
(44, 0.015)
(45, 0.013)
(46, 0.013)
(47, 0.011)
(48, 0.005)
(49, 0.009)
(50, 0.002)
(51, 0.001)
(52, 0.003)
(53, 0.002)
(54, 0.002)
(55, 0.001)
(56, 0.001)
(57, 0.001)
(58, 0.001)
(59, 0.001)
(60, 0.001)
(61, 0.001)
(62, 0.001)
(63, 0.002)
(64, 0.001)
(65, 0.001)
(66, 0.001)
(67, 0.001)
(68, 0.001)
(69, 0.001)
(70, 0.002)
(71, 0.001)
(72, 0.001)
(73, 0.001)
(74, 0.001)
(75, 0.001)
(76, 0.001)
(77, 0.001)
(78, 0.001)
(79, 0.001)
(80, 0.001)
(81, 0.001)
(82, 0.001)
(83, 0.001)
(84, 0.001)
(85, 0.001)
(86, 0.001)
(87, 0.001)
(88, 0.001)
(89, 0.001)
(90, 0.001)
(91, 0.001)
(92, 0.001)
(93, 0.001)
(94, 0.001)
(95, 0.001)
(96, 0.001)
(97, 0.001)
(98, 0.001)
(99, 0.001)
};
    
\addplot[
    color=cyan,
    ]
    coordinates { % 250-step PGD
(0, 2.679)
(1, 1.907)
(2, 1.671)
(3, 1.484)
(4, 1.309)
(5, 1.162)
(6, 1.011)
(7, 0.843)
(8, 0.367)
(9, 0.142)
(10, 0.062)
(11, 0.034)
(12, 0.035)
(13, 0.012)
(14, 0.031)
(15, 0.017)
(16, 0.010)
(17, 0.004)
(18, 0.006)
(19, 0.006)
(20, 0.005)
(21, 0.004)
(22, 0.008)
(23, 0.011)
(24, 0.007)
(25, 0.006)
(26, 0.005)
(27, 0.008)
(28, 0.004)
(29, 0.003)
(30, 0.005)
(31, 0.007)
(32, 0.002)
(33, 0.004)
(34, 0.006)
(35, 0.005)
(36, 0.010)
(37, 0.001)
(38, 0.005)
(39, 0.008)
(40, 0.013)
(41, 0.003)
(42, 0.010)
(43, 0.002)
(44, 0.003)
(45, 0.005)
(46, 0.013)
(47, 0.004)
(48, 0.008)
(49, 0.002)
(50, 0.002)
(51, 0.001)
(52, 0.001)
(53, 0.001)
(54, 0.001)
(55, 0.001)
(56, 0.001)
(57, 0.001)
(58, 0.001)
(59, 0.001)
(60, 0.001)
(61, 0.001)
(62, 0.001)
(63, 0.001)
(64, 0.001)
(65, 0.001)
(66, 0.001)
(67, 0.001)
(68, 0.001)
(69, 0.001)
(70, 0.001)
(71, 0.001)
(72, 0.001)
(73, 0.001)
(74, 0.001)
(75, 0.001)
(76, 0.001)
(77, 0.001)
(78, 0.001)
(79, 0.001)
(80, 0.001)
(81, 0.001)
(82, 0.001)
(83, 0.001)
(84, 0.001)
(85, 0.001)
(86, 0.001)
(87, 0.001)
(88, 0.001)
(89, 0.001)
(90, 0.001)
(91, 0.001)
(92, 0.001)
(93, 0.001)
(94, 0.001)
(95, 0.001)
(96, 0.001)
(97, 0.001)
(98, 0.001)
(99, 0.001)
};
    
\legend{Clean, RN, CU, P250}
    
\end{axis}
\end{tikzpicture}
        }
        % \caption{}
        % \label{subfig:selected-loss-in-aggregate}
    \end{subfigure}
    \begin{subfigure}[b]{0.28\textwidth}
        \centering
        \resizebox{\textwidth}{!}{
            \begin{tikzpicture}
\begin{axis}[
    xlabel={Epoch},
    ylabel={Train Loss},
    xmin=0, xmax=20,
    ymin=0, ymax=2.5,
    xtick={0,5,10,15,20},
    ytick={0,0.5,1,1.5,2,2.5},
    legend pos=north east,
    legend style={nodes={scale=0.7, transform shape}},
    ymajorgrids=true,
    grid style=dashed,
    colormap name=bright,
    cycle list={[of colormap]},
    every axis plot/.append style={mark=none,very thick},
]

\addplot
    coordinates { % R1
(0, 1.567)
(1, 1.231)
(2, 0.981)
(3, 0.809)
(4, 0.659)
(5, 0.639)
(6, 0.567)
(7, 0.494)
(8, 0.461)
(9, 0.421)
(10, 0.419)
(11, 0.300)
(12, 0.334)
(13, 0.321)
(14, 0.325)
(15, 0.292)
(16, 0.231)
(17, 0.247)
(18, 0.205)
(19, 0.180)
(20, 0.163)
(21, 0.193)
(22, 0.123)
(23, 0.170)
(24, 0.126)
(25, 0.118)
(26, 0.092)
(27, 0.113)
(28, 0.139)
(29, 0.104)
(30, 0.101)
(31, 0.115)
(32, 0.105)
(33, 0.103)
(34, 0.088)
(35, 0.095)
(36, 0.089)
(37, 0.097)
(38, 0.076)
(39, 0.118)
(40, 0.084)
(41, 0.088)
(42, 0.062)
(43, 0.075)
(44, 0.100)
(45, 0.066)
(46, 0.080)
(47, 0.075)
(48, 0.069)
(49, 0.065)
(50, 0.031)
(51, 0.017)
(52, 0.016)
(53, 0.013)
(54, 0.011)
(55, 0.012)
(56, 0.012)
(57, 0.011)
(58, 0.005)
(59, 0.005)
(60, 0.004)
(61, 0.004)
(62, 0.004)
(63, 0.004)
(64, 0.004)
(65, 0.004)
(66, 0.003)
(67, 0.004)
(68, 0.003)
(69, 0.003)
(70, 0.002)
(71, 0.004)
(72, 0.003)
(73, 0.003)
(74, 0.002)
(75, 0.002)
(76, 0.002)
(77, 0.003)
(78, 0.003)
(79, 0.002)
(80, 0.002)
(81, 0.002)
(82, 0.002)
(83, 0.002)
(84, 0.002)
(85, 0.001)
(86, 0.002)
(87, 0.002)
(88, 0.001)
(89, 0.003)
(90, 0.002)
(91, 0.002)
(92, 0.002)
(93, 0.003)
(94, 0.002)
(95, 0.002)
(96, 0.002)
(97, 0.002)
(98, 0.001)
(99, 0.002)
};

\addplot
    coordinates { % R2
(0, 1.607)
(1, 0.953)
(2, 0.583)
(3, 0.406)
(4, 0.255)
(5, 0.218)
(6, 0.193)
(7, 0.160)
(8, 0.151)
(9, 0.136)
(10, 0.115)
(11, 0.099)
(12, 0.079)
(13, 0.074)
(14, 0.048)
(15, 0.060)
(16, 0.049)
(17, 0.057)
(18, 0.044)
(19, 0.046)
(20, 0.048)
(21, 0.043)
(22, 0.043)
(23, 0.049)
(24, 0.028)
(25, 0.032)
(26, 0.021)
(27, 0.033)
(28, 0.033)
(29, 0.032)
(30, 0.027)
(31, 0.020)
(32, 0.017)
(33, 0.063)
(34, 0.038)
(35, 0.018)
(36, 0.019)
(37, 0.014)
(38, 0.021)
(39, 0.045)
(40, 0.016)
(41, 0.022)
(42, 0.026)
(43, 0.014)
(44, 0.020)
(45, 0.037)
(46, 0.016)
(47, 0.024)
(48, 0.018)
(49, 0.388)
(50, 0.020)
(51, 0.008)
(52, 0.007)
(53, 0.003)
(54, 0.004)
(55, 0.002)
(56, 0.004)
(57, 0.002)
(58, 0.003)
(59, 0.002)
(60, 0.001)
(61, 0.001)
(62, 0.002)
(63, 0.001)
(64, 0.001)
(65, 0.002)
(66, 0.001)
(67, 0.002)
(68, 0.001)
(69, 0.001)
(70, 0.001)
(71, 0.001)
(72, 0.001)
(73, 0.001)
(74, 0.003)
(75, 0.002)
(76, 0.001)
(77, 0.001)
(78, 0.001)
(79, 0.001)
(80, 0.001)
(81, 0.001)
(82, 0.002)
(83, 0.001)
(84, 0.001)
(85, 0.002)
(86, 0.001)
(87, 0.001)
(88, 0.001)
(89, 0.001)
(90, 0.001)
(91, 0.001)
(92, 0.001)
(93, 0.001)
(94, 0.001)
(95, 0.001)
(96, 0.001)
(97, 0.001)
(98, 0.001)
(99, 0.001)
};

\addplot
    coordinates { % R4
(0, 2.209)
(1, 1.194)
(2, 0.592)
(3, 0.373)
(4, 0.268)
(5, 0.231)
(6, 0.140)
(7, 0.107)
(8, 0.106)
(9, 0.118)
(10, 0.072)
(11, 0.077)
(12, 0.061)
(13, 0.069)
(14, 0.051)
(15, 0.049)
(16, 0.037)
(17, 0.044)
(18, 0.044)
(19, 0.035)
(20, 0.032)
(21, 0.039)
(22, 0.026)
(23, 0.019)
(24, 0.026)
(25, 0.014)
(26, 0.021)
(27, 0.035)
(28, 0.023)
(29, 0.017)
(30, 0.020)
(31, 0.014)
(32, 0.020)
(33, 0.017)
(34, 0.016)
(35, 0.016)
(36, 0.008)
(37, 0.010)
(38, 0.010)
(39, 0.019)
(40, 0.019)
(41, 0.009)
(42, 0.016)
(43, 0.011)
(44, 0.018)
(45, 0.016)
(46, 0.008)
(47, 0.010)
(48, 0.014)
(49, 0.014)
(50, 0.008)
(51, 0.002)
(52, 0.002)
(53, 0.002)
(54, 0.001)
(55, 0.002)
(56, 0.001)
(57, 0.001)
(58, 0.001)
(59, 0.002)
(60, 0.001)
(61, 0.001)
(62, 0.001)
(63, 0.001)
(64, 0.001)
(65, 0.002)
(66, 0.001)
(67, 0.001)
(68, 0.001)
(69, 0.002)
(70, 0.001)
(71, 0.001)
(72, 0.001)
(73, 0.001)
(74, 0.001)
(75, 0.001)
(76, 0.001)
(77, 0.001)
(78, 0.001)
(79, 0.001)
(80, 0.001)
(81, 0.002)
(82, 0.002)
(83, 0.001)
(84, 0.002)
(85, 0.001)
(86, 0.002)
(87, 0.001)
(88, 0.002)
(89, 0.002)
(90, 0.001)
(91, 0.002)
(92, 0.002)
(93, 0.002)
(94, 0.001)
(95, 0.001)
(96, 0.001)
(97, 0.002)
(98, 0.001)
(99, 0.002)
};

\addplot
    coordinates { % R16
(0, 1.693)
(1, 0.599)
(2, 0.073)
(3, 0.051)
(4, 0.043)
(5, 0.024)
(6, 0.012)
(7, 0.016)
(8, 0.005)
(9, 0.009)
(10, 0.008)
(11, 0.006)
(12, 0.004)
(13, 0.008)
(14, 0.006)
(15, 0.007)
(16, 0.005)
(17, 0.006)
(18, 0.004)
(19, 0.003)
(20, 0.008)
(21, 0.004)
(22, 0.004)
(23, 0.004)
(24, 0.006)
(25, 0.004)
(26, 0.009)
(27, 0.010)
(28, 0.005)
(29, 0.002)
(30, 0.004)
(31, 0.010)
(32, 0.004)
(33, 0.005)
(34, 0.003)
(35, 0.004)
(36, 0.002)
(37, 0.001)
(38, 0.001)
(39, 0.003)
(40, 0.008)
(41, 0.002)
(42, 0.005)
(43, 0.002)
(44, 0.003)
(45, 0.003)
(46, 0.001)
(47, 0.007)
(48, 0.003)
(49, 0.001)
(50, 0.001)
(51, 0.001)
(52, 0.001)
(53, 0.001)
(54, 0.001)
(55, 0.001)
(56, 0.001)
(57, 0.001)
(58, 0.001)
(59, 0.001)
(60, 0.001)
(61, 0.001)
(62, 0.001)
(63, 0.001)
(64, 0.001)
(65, 0.001)
(66, 0.001)
(67, 0.001)
(68, 0.001)
(69, 0.001)
(70, 0.001)
(71, 0.001)
(72, 0.001)
(73, 0.001)
(74, 0.001)
(75, 0.001)
(76, 0.001)
(77, 0.001)
(78, 0.001)
(79, 0.001)
(80, 0.001)
(81, 0.001)
(82, 0.001)
(83, 0.001)
(84, 0.001)
(85, 0.001)
(86, 0.001)
(87, 0.001)
(88, 0.001)
(89, 0.001)
(90, 0.001)
(91, 0.001)
(92, 0.001)
(93, 0.001)
(94, 0.001)
(95, 0.001)
(96, 0.001)
(97, 0.001)
(98, 0.001)
(99, 0.001)
};

\addplot
    coordinates { %R64
(0, 1.853)
(1, 1.518)
(2, 0.441)
(3, 0.115)
(4, 0.035)
(5, 0.020)
(6, 0.007)
(7, 0.010)
(8, 0.009)
(9, 0.004)
(10, 0.002)
(11, 0.003)
(12, 0.003)
(13, 0.003)
(14, 0.001)
(15, 0.001)
(16, 0.004)
(17, 0.002)
(18, 0.006)
(19, 0.002)
(20, 0.003)
(21, 0.009)
(22, 0.008)
(23, 0.003)
(24, 0.002)
(25, 0.002)
(26, 0.003)
(27, 0.001)
(28, 0.001)
(29, 0.001)
(30, 0.005)
(31, 0.003)
(32, 0.007)
(33, 0.013)
(34, 0.004)
(35, 0.002)
(36, 0.001)
(37, 0.005)
(38, 0.002)
(39, 0.001)
(40, 0.001)
(41, 0.001)
(42, 0.003)
(43, 0.011)
(44, 0.006)
(45, 0.003)
(46, 0.007)
(47, 0.002)
(48, 0.002)
(49, 0.005)
(50, 0.001)
(51, 0.001)
(52, 0.001)
(53, 0.001)
(54, 0.001)
(55, 0.001)
(56, 0.001)
(57, 0.001)
(58, 0.001)
(59, 0.001)
(60, 0.001)
(61, 0.001)
(62, 0.001)
(63, 0.001)
(64, 0.001)
(65, 0.001)
(66, 0.001)
(67, 0.001)
(68, 0.001)
(69, 0.001)
(70, 0.001)
(71, 0.001)
(72, 0.001)
(73, 0.001)
(74, 0.001)
(75, 0.001)
(76, 0.001)
(77, 0.001)
(78, 0.001)
(79, 0.001)
(80, 0.001)
(81, 0.001)
(82, 0.001)
(83, 0.001)
(84, 0.001)
(85, 0.001)
(86, 0.001)
(87, 0.001)
(88, 0.001)
(89, 0.001)
(90, 0.001)
(91, 0.001)
(92, 0.001)
(93, 0.001)
(94, 0.001)
(95, 0.001)
(96, 0.001)
(97, 0.001)
(98, 0.001)
(99, 0.001)
    };

\addplot
    coordinates { % R128
(0, 1.802)
(1, 1.537)
(2, 1.207)
(3, 0.480)
(4, 0.093)
(5, 0.030)
(6, 0.022)
(7, 0.013)
(8, 0.008)
(9, 0.006)
(10, 0.007)
(11, 0.002)
(12, 0.002)
(13, 0.002)
(14, 0.002)
(15, 0.001)
(16, 0.001)
(17, 0.001)
(18, 0.001)
(19, 0.001)
(20, 0.001)
(21, 0.001)
(22, 0.002)
(23, 0.009)
(24, 0.004)
(25, 0.006)
(26, 0.002)
(27, 0.001)
(28, 0.001)
(29, 0.002)
(30, 0.001)
(31, 0.001)
(32, 0.003)
(33, 0.002)
(34, 0.001)
(35, 0.005)
(36, 0.005)
(37, 0.008)
(38, 0.007)
(39, 0.001)
(40, 0.004)
(41, 0.001)
(42, 0.002)
(43, 0.001)
(44, 0.002)
(45, 0.001)
(46, 0.001)
(47, 0.002)
(48, 0.003)
(49, 0.003)
(50, 0.001)
(51, 0.001)
(52, 0.001)
(53, 0.001)
(54, 0.001)
(55, 0.001)
(56, 0.001)
(57, 0.001)
(58, 0.001)
(59, 0.001)
(60, 0.001)
(61, 0.001)
(62, 0.001)
(63, 0.001)
(64, 0.001)
(65, 0.001)
(66, 0.001)
(67, 0.001)
(68, 0.001)
(69, 0.001)
(70, 0.001)
(71, 0.001)
(72, 0.001)
(73, 0.001)
(74, 0.001)
(75, 0.001)
(76, 0.001)
(77, 0.001)
(78, 0.001)
(79, 0.001)
(80, 0.001)
(81, 0.001)
(82, 0.001)
(83, 0.001)
(84, 0.001)
(85, 0.001)
(86, 0.001)
(87, 0.001)
(88, 0.001)
(89, 0.001)
(90, 0.001)
(91, 0.001)
(92, 0.001)
(93, 0.001)
(94, 0.001)
(95, 0.001)
(96, 0.001)
(97, 0.001)
(98, 0.001)
(99, 0.001)
    };

\legend{R1, R2, R4, R16, R64, R128}
    
\end{axis}
\end{tikzpicture}
        }
        % \caption{}
        % \label{subfig:regions-loss-in-aggregate}
    \end{subfigure}
    \begin{subfigure}[b]{0.28\textwidth}
        \centering
        \resizebox{\textwidth}{!}{
            \begin{tikzpicture}
\begin{axis}[
    xlabel={Epoch},
    ylabel={Train Loss},
    xmin=0, xmax=20,
    ymin=0, ymax=2.5,
    xtick={0,5,10,15,20},
    ytick={0,0.5,1,1.5,2,2.5},
    legend pos=north east,
    legend style={nodes={scale=0.7, transform shape}},
    ymajorgrids=true,
    grid style=dashed,
    colormap name=bright,
    cycle list={[of colormap]},
    every axis plot/.append style={mark=none,very thick},
]

\addplot
    coordinates { % LF2
(0, 2.051)
(1, 1.798)
(2, 1.406)
(3, 0.876)
(4, 0.521)
(5, 0.483)
(6, 0.396)
(7, 0.324)
(8, 0.291)
(9, 0.216)
(10, 0.269)
(11, 0.198)
(12, 0.202)
(13, 0.178)
(14, 0.140)
(15, 0.149)
(16, 0.162)
(17, 0.130)
(18, 0.136)
(19, 0.116)
(20, 0.119)
(21, 0.093)
(22, 0.105)
(23, 0.076)
(24, 0.083)
(25, 0.106)
(26, 0.079)
(27, 0.070)
(28, 0.105)
(29, 0.091)
(30, 0.064)
(31, 0.104)
(32, 0.094)
(33, 0.041)
(34, 0.082)
(35, 0.073)
(36, 0.068)
(37, 0.061)
(38, 0.068)
(39, 0.065)
(40, 0.069)
(41, 0.066)
(42, 0.071)
(43, 0.057)
(44, 0.087)
(45, 0.082)
(46, 0.081)
(47, 0.052)
(48, 0.057)
(49, 0.041)
(50, 0.013)
(51, 0.016)
(52, 0.006)
(53, 0.008)
(54, 0.004)
(55, 0.007)
(56, 0.005)
(57, 0.004)
(58, 0.004)
(59, 0.004)
(60, 0.003)
(61, 0.004)
(62, 0.004)
(63, 0.003)
(64, 0.003)
(65, 0.004)
(66, 0.003)
(67, 0.003)
(68, 0.002)
(69, 0.003)
(70, 0.003)
(71, 0.003)
(72, 0.003)
(73, 0.003)
(74, 0.004)
(75, 0.003)
(76, 0.002)
(77, 0.002)
(78, 0.002)
(79, 0.002)
(80, 0.002)
(81, 0.002)
(82, 0.003)
(83, 0.003)
(84, 0.003)
(85, 0.002)
(86, 0.002)
(87, 0.002)
(88, 0.002)
(89, 0.002)
(90, 0.002)
(91, 0.002)
(92, 0.003)
(93, 0.002)
(94, 0.003)
(95, 0.002)
(96, 0.002)
(97, 0.002)
(98, 0.002)
(99, 0.002)
    };
    
\addplot
    coordinates { %LF3
(0, 1.685)
(1, 0.981)
(2, 0.428)
(3, 0.277)
(4, 0.236)
(5, 0.165)
(6, 0.163)
(7, 0.146)
(8, 0.117)
(9, 0.079)
(10, 0.068)
(11, 0.060)
(12, 0.077)
(13, 0.064)
(14, 0.069)
(15, 0.066)
(16, 0.053)
(17, 0.071)
(18, 0.047)
(19, 0.041)
(20, 0.055)
(21, 0.058)
(22, 0.048)
(23, 0.052)
(24, 0.028)
(25, 0.036)
(26, 0.044)
(27, 0.034)
(28, 0.042)
(29, 0.030)
(30, 0.023)
(31, 0.037)
(32, 0.040)
(33, 0.038)
(34, 0.033)
(35, 0.030)
(36, 0.026)
(37, 0.027)
(38, 0.043)
(39, 0.043)
(40, 0.040)
(41, 0.028)
(42, 0.021)
(43, 0.027)
(44, 0.018)
(45, 0.048)
(46, 0.039)
(47, 0.032)
(48, 0.026)
(49, 0.032)
(50, 0.014)
(51, 0.006)
(52, 0.002)
(53, 0.005)
(54, 0.002)
(55, 0.003)
(56, 0.002)
(57, 0.002)
(58, 0.002)
(59, 0.003)
(60, 0.002)
(61, 0.003)
(62, 0.002)
(63, 0.002)
(64, 0.002)
(65, 0.002)
(66, 0.002)
(67, 0.002)
(68, 0.003)
(69, 0.002)
(70, 0.002)
(71, 0.002)
(72, 0.002)
(73, 0.002)
(74, 0.002)
(75, 0.002)
(76, 0.001)
(77, 0.002)
(78, 0.002)
(79, 0.002)
(80, 0.002)
(81, 0.002)
(82, 0.002)
(83, 0.002)
(84, 0.002)
(85, 0.002)
(86, 0.002)
(87, 0.002)
(88, 0.002)
(89, 0.002)
(90, 0.002)
(91, 0.002)
(92, 0.002)
(93, 0.002)
(94, 0.002)
(95, 0.002)
(96, 0.002)
(97, 0.002)
(98, 0.002)
(99, 0.002)
    };
    
\addplot
    coordinates { % LF4
(0, 1.833)
(1, 1.619)
(2, 1.190)
(3, 0.763)
(4, 0.366)
(5, 0.270)
(6, 0.202)
(7, 0.145)
(8, 0.143)
(9, 0.118)
(10, 0.119)
(11, 0.100)
(12, 0.061)
(13, 0.079)
(14, 0.055)
(15, 0.069)
(16, 0.057)
(17, 0.050)
(18, 0.035)
(19, 0.041)
(20, 0.037)
(21, 0.028)
(22, 0.046)
(23, 0.031)
(24, 0.038)
(25, 0.032)
(26, 0.028)
(27, 0.030)
(28, 0.032)
(29, 0.026)
(30, 0.023)
(31, 0.033)
(32, 0.016)
(33, 0.026)
(34, 0.035)
(35, 0.023)
(36, 0.037)
(37, 0.033)
(38, 0.030)
(39, 0.023)
(40, 0.027)
(41, 0.021)
(42, 0.041)
(43, 0.018)
(44, 0.023)
(45, 0.039)
(46, 0.026)
(47, 0.037)
(48, 0.027)
(49, 0.026)
(50, 0.014)
(51, 0.004)
(52, 0.002)
(53, 0.003)
(54, 0.002)
(55, 0.002)
(56, 0.006)
(57, 0.001)
(58, 0.001)
(59, 0.002)
(60, 0.001)
(61, 0.002)
(62, 0.002)
(63, 0.002)
(64, 0.002)
(65, 0.001)
(66, 0.002)
(67, 0.001)
(68, 0.001)
(69, 0.002)
(70, 0.002)
(71, 0.002)
(72, 0.002)
(73, 0.002)
(74, 0.002)
(75, 0.001)
(76, 0.002)
(77, 0.002)
(78, 0.002)
(79, 0.002)
(80, 0.002)
(81, 0.002)
(82, 0.002)
(83, 0.002)
(84, 0.002)
(85, 0.002)
(86, 0.002)
(87, 0.002)
(88, 0.002)
(89, 0.002)
(90, 0.002)
(91, 0.002)
(92, 0.002)
(93, 0.002)
(94, 0.002)
(95, 0.002)
(96, 0.002)
(97, 0.002)
(98, 0.002)
(99, 0.002)
};
    
\addplot
    coordinates { % LF5
(0, 2.245)
(1, 1.837)
(2, 1.563)
(3, 1.422)
(4, 1.054)
(5, 0.682)
(6, 0.303)
(7, 0.232)
(8, 0.158)
(9, 0.118)
(10, 0.122)
(11, 0.123)
(12, 0.078)
(13, 0.054)
(14, 0.066)
(15, 0.057)
(16, 0.056)
(17, 0.061)
(18, 0.051)
(19, 0.053)
(20, 0.032)
(21, 0.045)
(22, 0.055)
(23, 0.028)
(24, 0.038)
(25, 0.031)
(26, 0.035)
(27, 0.021)
(28, 0.031)
(29, 0.033)
(30, 0.028)
(31, 0.019)
(32, 0.021)
(33, 0.028)
(34, 0.026)
(35, 0.026)
(36, 0.037)
(37, 0.020)
(38, 0.018)
(39, 0.035)
(40, 0.027)
(41, 0.023)
(42, 0.019)
(43, 0.014)
(44, 0.020)
(45, 0.019)
(46, 0.014)
(47, 0.030)
(48, 0.030)
(49, 0.017)
(50, 0.009)
(51, 0.003)
(52, 0.003)
(53, 0.002)
(54, 0.002)
(55, 0.002)
(56, 0.002)
(57, 0.002)
(58, 0.002)
(59, 0.001)
(60, 0.002)
(61, 0.004)
(62, 0.002)
(63, 0.002)
(64, 0.002)
(65, 0.002)
(66, 0.001)
(67, 0.002)
(68, 0.002)
(69, 0.002)
(70, 0.002)
(71, 0.002)
(72, 0.002)
(73, 0.002)
(74, 0.002)
(75, 0.003)
(76, 0.002)
(77, 0.002)
(78, 0.002)
(79, 0.002)
(80, 0.002)
(81, 0.002)
(82, 0.002)
(83, 0.002)
(84, 0.002)
(85, 0.002)
(86, 0.002)
(87, 0.002)
(88, 0.002)
(89, 0.002)
(90, 0.002)
(91, 0.002)
(92, 0.002)
(93, 0.002)
(94, 0.002)
(95, 0.002)
(96, 0.002)
(97, 0.002)
(98, 0.002)
(99, 0.002)
};

\addplot
    coordinates { % LF6
(0, 1.981)
(1, 1.604)
(2, 1.350)
(3, 1.133)
(4, 0.870)
(5, 0.515)
(6, 0.229)
(7, 0.105)
(8, 0.101)
(9, 0.079)
(10, 0.045)
(11, 0.042)
(12, 0.030)
(13, 0.041)
(14, 0.038)
(15, 0.024)
(16, 0.019)
(17, 0.024)
(18, 0.015)
(19, 0.027)
(20, 0.020)
(21, 0.021)
(22, 0.020)
(23, 0.029)
(24, 0.025)
(25, 0.010)
(26, 0.031)
(27, 0.017)
(28, 0.021)
(29, 0.023)
(30, 0.015)
(31, 0.011)
(32, 0.021)
(33, 0.027)
(34, 0.016)
(35, 0.012)
(36, 0.028)
(37, 0.013)
(38, 0.019)
(39, 0.010)
(40, 0.014)
(41, 0.019)
(42, 0.013)
(43, 0.020)
(44, 0.018)
(45, 0.011)
(46, 0.022)
(47, 0.029)
(48, 0.017)
(49, 0.013)
(50, 0.003)
(51, 0.002)
(52, 0.002)
(53, 0.002)
(54, 0.001)
(55, 0.002)
(56, 0.001)
(57, 0.001)
(58, 0.001)
(59, 0.001)
(60, 0.002)
(61, 0.001)
(62, 0.001)
(63, 0.002)
(64, 0.001)
(65, 0.001)
(66, 0.001)
(67, 0.002)
(68, 0.001)
(69, 0.002)
(70, 0.001)
(71, 0.001)
(72, 0.002)
(73, 0.002)
(74, 0.002)
(75, 0.002)
(76, 0.002)
(77, 0.002)
(78, 0.002)
(79, 0.001)
(80, 0.001)
(81, 0.002)
(82, 0.002)
(83, 0.002)
(84, 0.001)
(85, 0.002)
(86, 0.002)
(87, 0.001)
(88, 0.002)
(89, 0.002)
(90, 0.001)
(91, 0.002)
(92, 0.002)
(93, 0.002)
(94, 0.002)
(95, 0.002)
(96, 0.002)
(97, 0.002)
(98, 0.002)
(99, 0.002)
};

\addplot
    coordinates { % LF7
(0, 2.129)
(1, 1.940)
(2, 1.707)
(3, 1.507)
(4, 1.414)
(5, 1.220)
(6, 0.986)
(7, 0.623)
(8, 0.371)
(9, 0.188)
(10, 0.107)
(11, 0.065)
(12, 0.061)
(13, 0.045)
(14, 0.028)
(15, 0.041)
(16, 0.028)
(17, 0.020)
(18, 0.017)
(19, 0.017)
(20, 0.025)
(21, 0.019)
(22, 0.011)
(23, 0.013)
(24, 0.015)
(25, 0.017)
(26, 0.012)
(27, 0.008)
(28, 0.011)
(29, 0.014)
(30, 0.016)
(31, 0.017)
(32, 0.015)
(33, 0.029)
(34, 0.012)
(35, 0.013)
(36, 0.015)
(37, 0.022)
(38, 0.016)
(39, 0.013)
(40, 0.017)
(41, 0.016)
(42, 0.007)
(43, 0.017)
(44, 0.019)
(45, 0.019)
(46, 0.011)
(47, 0.017)
(48, 0.010)
(49, 0.020)
(50, 0.003)
(51, 0.003)
(52, 0.003)
(53, 0.001)
(54, 0.003)
(55, 0.002)
(56, 0.002)
(57, 0.001)
(58, 0.001)
(59, 0.002)
(60, 0.001)
(61, 0.001)
(62, 0.001)
(63, 0.001)
(64, 0.001)
(65, 0.001)
(66, 0.001)
(67, 0.002)
(68, 0.001)
(69, 0.002)
(70, 0.001)
(71, 0.002)
(72, 0.002)
(73, 0.001)
(74, 0.001)
(75, 0.002)
(76, 0.002)
(77, 0.002)
(78, 0.002)
(79, 0.002)
(80, 0.002)
(81, 0.002)
(82, 0.002)
(83, 0.001)
(84, 0.002)
(85, 0.001)
(86, 0.002)
(87, 0.002)
(88, 0.002)
(89, 0.001)
(90, 0.002)
(91, 0.002)
(92, 0.002)
(93, 0.002)
(94, 0.002)
(95, 0.002)
(96, 0.002)
(97, 0.002)
(98, 0.002)
(99, 0.002)
};

\addplot
    coordinates { % LF8
(0, 1.924)
(1, 1.604)
(2, 1.287)
(3, 1.097)
(4, 0.990)
(5, 0.853)
(6, 0.691)
(7, 0.606)
(8, 0.530)
(9, 0.449)
(10, 0.362)
(11, 0.288)
(12, 0.195)
(13, 0.112)
(14, 0.077)
(15, 0.068)
(16, 0.034)
(17, 0.024)
(18, 0.023)
(19, 0.021)
(20, 0.029)
(21, 0.011)
(22, 0.012)
(23, 0.012)
(24, 0.025)
(25, 0.004)
(26, 0.015)
(27, 0.009)
(28, 0.011)
(29, 0.011)
(30, 0.015)
(31, 0.012)
(32, 0.021)
(33, 0.020)
(34, 0.011)
(35, 0.015)
(36, 0.015)
(37, 0.013)
(38, 0.014)
(39, 0.014)
(40, 0.012)
(41, 0.013)
(42, 0.019)
(43, 0.008)
(44, 0.023)
(45, 0.015)
(46, 0.016)
(47, 0.019)
(48, 0.013)
(49, 0.011)
(50, 0.008)
(51, 0.003)
(52, 0.001)
(53, 0.001)
(54, 0.001)
(55, 0.001)
(56, 0.002)
(57, 0.001)
(58, 0.001)
(59, 0.001)
(60, 0.001)
(61, 0.001)
(62, 0.001)
(63, 0.001)
(64, 0.001)
(65, 0.001)
(66, 0.001)
(67, 0.001)
(68, 0.001)
(69, 0.001)
(70, 0.001)
(71, 0.001)
(72, 0.001)
(73, 0.001)
(74, 0.001)
(75, 0.001)
(76, 0.001)
(77, 0.001)
(78, 0.001)
(79, 0.001)
(80, 0.002)
(81, 0.001)
(82, 0.001)
(83, 0.001)
(84, 0.001)
(85, 0.001)
(86, 0.001)
(87, 0.001)
(88, 0.001)
(89, 0.001)
(90, 0.001)
(91, 0.001)
(92, 0.002)
(93, 0.001)
(94, 0.001)
(95, 0.001)
(96, 0.001)
(97, 0.001)
(98, 0.001)
(99, 0.001)
};
    
\legend{LF2, LF3, LF4, LF5, LF6, LF7, LF8}
    
\end{axis}
\end{tikzpicture}
        }
        % \caption{}
        % \label{subfig:low-freq-loss-in-aggregate}
    \end{subfigure}
    
    \begin{subfigure}[b]{0.28\textwidth}
        \centering
        \resizebox{\textwidth}{!}{
            \begin{tikzpicture}
\begin{axis}[
    xlabel={Epoch},
    ylabel={Test Accuracy},
    xmin=0, xmax=100,
    ymin=0, ymax=100,
    xtick={0,20,40,60,80,100},
    ytick={0,20,40,60,80,100},
    legend pos=north east,
    legend style={nodes={scale=0.7, transform shape}},
    ymajorgrids=true,
    grid style=dashed,
    every axis plot/.append style={very thick}
]

\addplot[
    color=green,
    ]
    coordinates { %Clean
(0, 36.780)
(1, 48.900)
(2, 51.040)
(3, 55.340)
(4, 59.850)
(5, 67.400)
(6, 73.140)
(7, 76.490)
(8, 70.970)
(9, 74.100)
(10, 78.000)
(11, 81.770)
(12, 81.400)
(13, 83.200)
(14, 82.330)
(15, 84.470)
(16, 84.360)
(17, 85.350)
(18, 86.250)
(19, 83.450)
(20, 84.290)
(21, 86.670)
(22, 85.830)
(23, 85.470)
(24, 87.830)
(25, 85.360)
(26, 87.840)
(27, 86.970)
(28, 86.810)
(29, 83.170)
(30, 87.190)
(31, 87.150)
(32, 88.410)
(33, 82.900)
(34, 84.880)
(35, 84.330)
(36, 85.990)
(37, 88.510)
(38, 87.630)
(39, 89.640)
(40, 89.290)
(41, 87.960)
(42, 87.000)
(43, 88.370)
(44, 87.810)
(45, 85.400)
(46, 85.340)
(47, 87.570)
(48, 87.430)
(49, 85.590)
(50, 93.100)
(51, 93.340)
(52, 93.470)
(53, 93.630)
(54, 93.620)
(55, 93.760)
(56, 93.640)
(57, 93.850)
(58, 93.750)
(59, 93.650)
(60, 93.730)
(61, 93.840)
(62, 93.790)
(63, 93.830)
(64, 93.910)
(65, 94.010)
(66, 94.050)
(67, 94.120)
(68, 94.000)
(69, 93.920)
(70, 93.810)
(71, 93.980)
(72, 94.130)
(73, 93.930)
(74, 93.940)
(75, 93.900)
(76, 94.100)
(77, 94.010)
(78, 93.920)
(79, 93.960)
(80, 94.130)
(81, 94.250)
(82, 94.060)
(83, 94.220)
(84, 94.020)
(85, 94.280)
(86, 93.940)
(87, 94.120)
(88, 94.150)
(89, 94.110)
(90, 94.150)
(91, 93.970)
(92, 94.230)
(93, 94.130)
(94, 94.050)
(95, 93.960)
(96, 94.070)
(97, 94.070)
(98, 94.100)
(99, 94.120)
    };
    
\addplot[
    color=red,
    ]
    coordinates { %Class-wise Random
(0, 35.420)
(1, 45.680)
(2, 53.880)
(3, 53.200)
(4, 64.430)
(5, 66.580)
(6, 57.920)
(7, 53.990)
(8, 44.120)
(9, 25.560)
(10, 27.960)
(11, 27.520)
(12, 21.930)
(13, 19.860)
(14, 23.330)
(15, 25.640)
(16, 20.890)
(17, 20.370)
(18, 19.030)
(19, 16.460)
(20, 16.760)
(21, 16.230)
(22, 16.920)
(23, 14.230)
(24, 17.620)
(25, 14.150)
(26, 11.120)
(27, 13.790)
(28, 14.430)
(29, 13.080)
(30, 13.550)
(31, 10.870)
(32, 10.560)
(33, 13.910)
(34, 16.010)
(35, 13.060)
(36, 10.390)
(37, 11.260)
(38, 10.880)
(39, 13.710)
(40, 10.320)
(41, 10.090)
(42, 11.230)
(43, 9.970)
(44, 12.020)
(45, 13.410)
(46, 11.890)
(47, 10.090)
(48, 15.450)
(49, 12.080)
(50, 12.830)
(51, 12.360)
(52, 12.610)
(53, 12.070)
(54, 12.090)
(55, 12.190)
(56, 11.590)
(57, 12.540)
(58, 11.700)
(59, 12.280)
(60, 11.990)
(61, 11.820)
(62, 11.960)
(63, 12.200)
(64, 12.010)
(65, 11.870)
(66, 12.640)
(67, 12.000)
(68, 12.950)
(69, 12.340)
(70, 12.240)
(71, 12.350)
(72, 11.470)
(73, 11.240)
(74, 11.350)
(75, 11.690)
(76, 12.110)
(77, 11.600)
(78, 12.640)
(79, 11.630)
(80, 12.110)
(81, 11.560)
(82, 11.650)
(83, 11.870)
(84, 12.440)
(85, 11.650)
(86, 11.570)
(87, 11.480)
(88, 11.690)
(89, 11.700)
(90, 11.960)
(91, 12.090)
(92, 11.760)
(93, 11.160)
(94, 11.240)
(95, 11.760)
(96, 12.440)
(97, 11.740)
(98, 12.370)
(99, 11.640)
    };
    
\addplot[
    color=blue,
    ]
    coordinates { % Class-wise Unlearnable
(0, 33.220)
(1, 23.280)
(2, 16.430)
(3, 17.440)
(4, 14.600)
(5, 19.130)
(6, 15.070)
(7, 17.540)
(8, 16.080)
(9, 16.760)
(10, 14.740)
(11, 20.810)
(12, 18.670)
(13, 16.280)
(14, 18.220)
(15, 17.300)
(16, 16.720)
(17, 16.890)
(18, 16.420)
(19, 15.840)
(20, 19.690)
(21, 17.960)
(22, 19.000)
(23, 16.060)
(24, 13.780)
(25, 16.540)
(26, 18.850)
(27, 16.840)
(28, 17.790)
(29, 18.890)
(30, 16.700)
(31, 15.780)
(32, 19.360)
(33, 18.870)
(34, 19.440)
(35, 18.480)
(36, 20.650)
(37, 15.050)
(38, 19.500)
(39, 20.480)
(40, 18.870)
(41, 19.390)
(42, 22.650)
(43, 21.230)
(44, 21.550)
(45, 21.850)
(46, 20.090)
(47, 17.010)
(48, 19.940)
(49, 18.620)
(50, 18.640)
(51, 19.540)
(52, 20.380)
(53, 20.490)
(54, 20.420)
(55, 20.320)
(56, 20.370)
(57, 20.460)
(58, 20.090)
(59, 20.720)
(60, 20.150)
(61, 20.240)
(62, 20.490)
(63, 19.950)
(64, 19.790)
(65, 19.710)
(66, 19.800)
(67, 20.010)
(68, 19.990)
(69, 19.740)
(70, 19.450)
(71, 19.900)
(72, 19.340)
(73, 19.570)
(74, 19.270)
(75, 19.510)
(76, 19.880)
(77, 20.070)
(78, 19.480)
(79, 19.700)
(80, 19.600)
(81, 19.720)
(82, 19.510)
(83, 19.490)
(84, 19.730)
(85, 19.390)
(86, 19.650)
(87, 19.710)
(88, 19.630)
(89, 19.420)
(90, 19.420)
(91, 19.470)
(92, 19.700)
(93, 19.710)
(94, 19.440)
(95, 19.620)
(96, 19.570)
(97, 19.150)
(98, 19.490)
(99, 19.190)
};
    
\addplot[
    color=cyan,
    ]
    coordinates { % 250-step PGD
(0, 26.380)
(1, 36.620)
(2, 44.770)
(3, 46.510)
(4, 53.610)
(5, 57.170)
(6, 57.300)
(7, 39.400)
(8, 26.670)
(9, 24.070)
(10, 24.440)
(11, 20.960)
(12, 20.700)
(13, 22.170)
(14, 14.310)
(15, 14.540)
(16, 19.640)
(17, 18.830)
(18, 17.990)
(19, 18.690)
(20, 18.670)
(21, 18.580)
(22, 13.390)
(23, 14.420)
(24, 14.580)
(25, 14.500)
(26, 13.760)
(27, 16.820)
(28, 13.020)
(29, 9.010)
(30, 14.110)
(31, 11.270)
(32, 16.540)
(33, 18.320)
(34, 11.110)
(35, 13.080)
(36, 15.500)
(37, 13.450)
(38, 7.900)
(39, 13.100)
(40, 12.990)
(41, 10.390)
(42, 13.450)
(43, 15.020)
(44, 8.810)
(45, 13.600)
(46, 12.690)
(47, 8.410)
(48, 9.930)
(49, 9.660)
(50, 8.100)
(51, 8.500)
(52, 8.500)
(53, 7.700)
(54, 8.300)
(55, 7.980)
(56, 7.800)
(57, 7.920)
(58, 8.450)
(59, 7.860)
(60, 8.490)
(61, 7.870)
(62, 8.150)
(63, 8.120)
(64, 8.100)
(65, 8.090)
(66, 8.100)
(67, 7.820)
(68, 7.950)
(69, 8.380)
(70, 8.030)
(71, 8.280)
(72, 7.980)
(73, 8.690)
(74, 8.160)
(75, 8.150)
(76, 8.200)
(77, 8.130)
(78, 8.030)
(79, 8.480)
(80, 8.260)
(81, 8.220)
(82, 8.360)
(83, 8.040)
(84, 8.130)
(85, 8.500)
(86, 8.340)
(87, 8.310)
(88, 8.300)
(89, 8.320)
(90, 7.770)
(91, 7.920)
(92, 7.940)
(93, 7.850)
(94, 8.590)
(95, 7.950)
(96, 8.000)
(97, 8.140)
(98, 8.770)
(99, 8.150)
};
    
\legend{Clean, RN, CU, P250}
    
\end{axis}
\end{tikzpicture}
        }
        % \caption{}
        % \label{}
    \end{subfigure}
    \begin{subfigure}[b]{0.28\textwidth}
        \centering
        \resizebox{\textwidth}{!}{
            \begin{tikzpicture}
\begin{axis}[
    xlabel={Epoch},
    ylabel={Test Accuracy},
    xmin=0, xmax=100,
    ymin=0, ymax=100,
    xtick={0,20,40,60,80,100},
    ytick={0,20,40,60,80,100},
    legend pos=north east,
    legend style={nodes={scale=0.7, transform shape}},
    ymajorgrids=true,
    grid style=dashed,
    colormap name=bright,
    cycle list={[of colormap]},
    every axis plot/.append style={mark=none,very thick},
]

\addplot
    coordinates { % R1
(0, 21.3700)
(1, 28.8900)
(2, 31.4700)
(3, 36.4100)
(4, 30.2300)
(5, 40.4400)
(6, 40.8800)
(7, 40.9500)
(8, 44.7500)
(9, 45.6600)
(10, 42.2600)
(11, 49.8200)
(12, 55.0300)
(13, 50.8500)
(14, 50.4800)
(15, 53.1900)
(16, 50.9300)
(17, 55.1200)
(18, 59.9100)
(19, 60.7500)
(20, 54.7600)
(21, 60.4100)
(22, 62.8000)
(23, 59.2800)
(24, 62.3400)
(25, 58.0200)
(26, 62.8500)
(27, 64.4300)
(28, 56.3100)
(29, 61.3100)
(30, 64.0400)
(31, 66.4300)
(32, 57.0600)
(33, 63.4900)
(34, 61.8200)
(35, 63.1600)
(36, 69.1100)
(37, 62.5200)
(38, 55.4200)
(39, 65.5600)
(40, 59.0500)
(41, 56.1900)
(42, 64.8500)
(43, 64.9700)
(44, 56.8800)
(45, 64.9800)
(46, 60.8900)
(47, 63.7300)
(48, 67.8700)
(49, 67.9500)
(50, 72.8700)
(51, 74.3600)
(52, 73.3900)
(53, 74.2900)
(54, 73.4500)
(55, 74.5200)
(56, 73.1000)
(57, 74.6100)
(58, 74.1300)
(59, 74.6000)
(60, 75.8300)
(61, 73.3700)
(62, 73.9000)
(63, 75.4700)
(64, 74.5000)
(65, 75.2400)
(66, 73.0300)
(67, 74.1500)
(68, 74.6800)
(69, 74.7200)
(70, 74.2000)
(71, 74.2100)
(72, 73.9200)
(73, 74.4200)
(74, 74.7000)
(75, 74.7100)
(76, 75.7700)
(77, 75.0200)
(78, 75.6700)
(79, 75.3400)
(80, 75.2400)
(81, 74.0500)
(82, 74.4000)
(83, 74.6800)
(84, 74.4000)
(85, 74.8700)
(86, 74.1200)
(87, 74.3800)
(88, 74.6500)
(89, 75.2100)
(90, 74.5500)
(91, 74.6300)
(92, 73.7100)
(93, 74.2100)
(94, 74.2500)
(95, 73.9900)
(96, 73.8300)
(97, 74.9700)
(98, 74.0900)
(99, 74.7900)
    };
    
\addplot
    coordinates { %R2
(0, 21.6500)
(1, 22.4100)
(2, 24.6800)
(3, 26.7200)
(4, 25.2400)
(5, 27.1300)
(6, 26.3200)
(7, 23.8600)
(8, 27.1900)
(9, 28.9700)
(10, 27.4400)
(11, 27.8700)
(12, 26.7100)
(13, 24.0500)
(14, 26.9400)
(15, 28.1100)
(16, 23.4800)
(17, 26.7300)
(18, 23.0300)
(19, 26.4800)
(20, 22.5500)
(21, 28.7800)
(22, 25.9300)
(23, 30.1800)
(24, 27.3000)
(25, 26.5700)
(26, 27.5000)
(27, 26.1400)
(28, 25.4400)
(29, 25.1100)
(30, 29.0200)
(31, 29.6300)
(32, 28.1700)
(33, 25.4100)
(34, 29.9100)
(35, 31.1900)
(36, 25.1900)
(37, 26.3400)
(38, 26.7600)
(39, 30.4800)
(40, 27.4000)
(41, 23.4600)
(42, 27.6300)
(43, 27.3900)
(44, 27.9400)
(45, 32.0200)
(46, 22.3700)
(47, 27.4500)
(48, 28.7700)
(49, 26.9400)
(50, 34.1500)
(51, 34.9600)
(52, 34.3400)
(53, 33.6200)
(54, 34.1500)
(55, 34.0400)
(56, 34.8700)
(57, 33.7700)
(58, 34.6300)
(59, 34.4700)
(60, 34.0200)
(61, 34.0600)
(62, 34.1400)
(63, 34.0200)
(64, 33.6900)
(65, 33.7700)
(66, 33.8300)
(67, 33.6800)
(68, 33.1400)
(69, 33.8600)
(70, 32.4000)
(71, 32.3700)
(72, 33.4000)
(73, 32.8900)
(74, 33.4300)
(75, 33.1400)
(76, 32.2300)
(77, 33.4500)
(78, 31.6100)
(79, 32.6700)
(80, 31.9400)
(81, 32.3700)
(82, 32.4800)
(83, 31.7200)
(84, 31.8500)
(85, 31.4400)
(86, 32.6600)
(87, 32.9600)
(88, 31.2100)
(89, 31.6300)
(90, 32.2000)
(91, 32.2200)
(92, 32.2400)
(93, 31.7800)
(94, 31.1600)
(95, 31.5800)
(96, 32.0700)
(97, 32.1800)
(98, 32.3900)
(99, 31.1400)
    };
    
\addplot
    coordinates { % R4
(0, 24.5100)
(1, 21.9200)
(2, 21.2300)
(3, 22.3000)
(4, 20.3400)
(5, 20.9400)
(6, 18.4100)
(7, 21.9100)
(8, 20.7500)
(9, 20.6000)
(10, 22.5700)
(11, 18.9400)
(12, 19.9400)
(13, 20.7900)
(14, 19.2300)
(15, 16.1100)
(16, 19.2300)
(17, 17.2200)
(18, 17.5300)
(19, 15.7300)
(20, 17.5700)
(21, 16.6200)
(22, 18.4900)
(23, 17.0200)
(24, 17.5000)
(25, 15.2500)
(26, 16.7100)
(27, 15.6300)
(28, 15.8200)
(29, 17.3300)
(30, 17.8900)
(31, 16.9400)
(32, 15.9100)
(33, 14.7200)
(34, 17.3800)
(35, 16.3100)
(36, 16.8600)
(37, 17.9600)
(38, 13.6700)
(39, 15.6700)
(40, 14.7100)
(41, 15.4600)
(42, 12.7000)
(43, 15.5800)
(44, 14.8900)
(45, 12.7600)
(46, 12.0200)
(47, 14.7100)
(48, 11.5200)
(49, 14.9700)
(50, 11.6600)
(51, 11.7300)
(52, 11.9100)
(53, 11.8200)
(54, 11.6900)
(55, 11.7600)
(56, 11.4000)
(57, 11.9500)
(58, 11.7900)
(59, 11.9200)
(60, 11.7000)
(61, 11.7200)
(62, 11.5800)
(63, 11.6900)
(64, 11.7500)
(65, 11.6900)
(66, 11.6000)
(67, 11.5600)
(68, 11.4700)
(69, 11.7000)
(70, 11.6500)
(71, 11.7100)
(72, 11.8200)
(73, 11.6000)
(74, 11.9800)
(75, 11.7400)
(76, 11.6800)
(77, 11.6200)
(78, 11.8800)
(79, 11.5200)
(80, 11.5800)
(81, 11.6100)
(82, 11.6900)
(83, 11.5800)
(84, 11.4500)
(85, 11.8100)
(86, 11.5500)
(87, 11.7500)
(88, 11.6900)
(89, 11.7700)
(90, 11.6800)
(91, 11.6000)
(92, 11.6000)
(93, 11.6700)
(94, 11.6300)
(95, 11.5600)
(96, 11.6900)
(97, 11.6000)
(98, 11.6200)
(99, 11.4500)
};

\addplot
    coordinates { % R16
(0, 22.0100)
(1, 19.8200)
(2, 16.9900)
(3, 21.8600)
(4, 21.0200)
(5, 24.5400)
(6, 19.7100)
(7, 19.4700)
(8, 22.3700)
(9, 19.4400)
(10, 18.6300)
(11, 22.8600)
(12, 21.4600)
(13, 22.5000)
(14, 21.1100)
(15, 21.6500)
(16, 22.8200)
(17, 21.7000)
(18, 24.3100)
(19, 21.3800)
(20, 21.3500)
(21, 19.8700)
(22, 22.1500)
(23, 23.7000)
(24, 20.9900)
(25, 20.5000)
(26, 22.9000)
(27, 19.0800)
(28, 18.7600)
(29, 19.9100)
(30, 21.2900)
(31, 19.4000)
(32, 16.3700)
(33, 19.2000)
(34, 16.8200)
(35, 15.6700)
(36, 20.3600)
(37, 17.0000)
(38, 14.9900)
(39, 19.0300)
(40, 17.8900)
(41, 22.7900)
(42, 22.5200)
(43, 15.5800)
(44, 14.3900)
(45, 15.5800)
(46, 17.8500)
(47, 16.9800)
(48, 16.3600)
(49, 16.5900)
(50, 21.4300)
(51, 21.7800)
(52, 21.3700)
(53, 21.9900)
(54, 22.1900)
(55, 21.8200)
(56, 22.1300)
(57, 21.9600)
(58, 22.0800)
(59, 21.4100)
(60, 22.0100)
(61, 21.8300)
(62, 21.7000)
(63, 21.4400)
(64, 21.3200)
(65, 21.2300)
(66, 21.7000)
(67, 21.4900)
(68, 21.1700)
(69, 21.3300)
(70, 21.2000)
(71, 21.2800)
(72, 20.7700)
(73, 21.2200)
(74, 20.9200)
(75, 20.8700)
(76, 20.6300)
(77, 21.1600)
(78, 20.6000)
(79, 21.3400)
(80, 21.1000)
(81, 20.8800)
(82, 21.9000)
(83, 21.4800)
(84, 20.9400)
(85, 21.1200)
(86, 20.5900)
(87, 21.2300)
(88, 21.1700)
(89, 21.4300)
(90, 21.1900)
(91, 21.2900)
(92, 21.1400)
(93, 21.0200)
(94, 19.8900)
(95, 20.7900)
(96, 21.9100)
(97, 21.5900)
(98, 21.1400)
(99, 20.3200)
};

\addplot
    coordinates { % R64
(0, 33.2100)
(1, 27.1100)
(2, 20.9700)
(3, 16.5600)
(4, 17.5100)
(5, 15.2200)
(6, 14.7400)
(7, 15.9600)
(8, 19.7800)
(9, 16.8700)
(10, 16.3200)
(11, 15.3100)
(12, 16.3200)
(13, 16.9500)
(14, 16.9800)
(15, 15.4600)
(16, 15.8200)
(17, 17.9100)
(18, 11.0700)
(19, 16.7300)
(20, 16.6000)
(21, 16.0100)
(22, 18.3600)
(23, 19.7900)
(24, 16.8800)
(25, 15.3400)
(26, 14.9900)
(27, 16.3500)
(28, 15.3700)
(29, 16.5400)
(30, 13.9000)
(31, 13.8500)
(32, 15.1900)
(33, 17.3500)
(34, 18.7300)
(35, 16.5200)
(36, 15.9500)
(37, 16.6000)
(38, 14.3700)
(39, 16.9200)
(40, 17.1000)
(41, 12.3100)
(42, 16.0500)
(43, 13.7500)
(44, 15.8600)
(45, 16.2400)
(46, 15.3400)
(47, 13.4200)
(48, 11.9700)
(49, 14.4300)
(50, 15.9000)
(51, 15.9500)
(52, 14.9800)
(53, 15.4700)
(54, 16.0600)
(55, 16.7300)
(56, 15.5700)
(57, 16.9300)
(58, 15.5600)
(59, 15.5400)
(60, 16.3800)
(61, 15.7400)
(62, 16.7800)
(63, 17.2900)
(64, 16.7900)
(65, 16.6600)
(66, 16.9100)
(67, 15.9500)
(68, 16.5500)
(69, 16.0100)
(70, 15.4800)
(71, 15.5100)
(72, 14.6100)
(73, 16.1800)
(74, 15.1700)
(75, 16.0000)
(76, 14.8500)
(77, 14.8900)
(78, 15.0200)
(79, 15.2200)
(80, 15.2400)
(81, 14.4000)
(82, 14.5800)
(83, 14.5600)
(84, 14.4000)
(85, 14.4300)
(86, 14.0700)
(87, 14.3300)
(88, 13.7000)
(89, 13.6300)
(90, 14.0300)
(91, 14.0500)
(92, 14.2100)
(93, 13.4700)
(94, 13.1600)
(95, 13.7000)
(96, 13.0300)
(97, 13.0700)
(98, 13.2600)
(99, 13.0000)
};

\addplot
    coordinates { % R128
(0, 38.1000)
(1, 44.2600)
(2, 42.9500)
(3, 28.8500)
(4, 23.0600)
(5, 25.8300)
(6, 24.9600)
(7, 23.7700)
(8, 24.3000)
(9, 26.0000)
(10, 26.9600)
(11, 25.3800)
(12, 26.1800)
(13, 25.4800)
(14, 24.4400)
(15, 26.3400)
(16, 28.8500)
(17, 24.6300)
(18, 27.7700)
(19, 28.4500)
(20, 27.0700)
(21, 27.5300)
(22, 21.0100)
(23, 19.6700)
(24, 21.6900)
(25, 18.4800)
(26, 20.2100)
(27, 18.6800)
(28, 20.0600)
(29, 21.4700)
(30, 17.0200)
(31, 15.5400)
(32, 14.3900)
(33, 13.9300)
(34, 16.0000)
(35, 17.3200)
(36, 13.1800)
(37, 14.0300)
(38, 14.7700)
(39, 15.2200)
(40, 15.0800)
(41, 12.6200)
(42, 13.1100)
(43, 11.0500)
(44, 14.9300)
(45, 12.7800)
(46, 13.3400)
(47, 12.8800)
(48, 10.2300)
(49, 12.2500)
(50, 14.0400)
(51, 13.2000)
(52, 13.7200)
(53, 13.4700)
(54, 13.5500)
(55, 13.6500)
(56, 13.3400)
(57, 13.4100)
(58, 13.0600)
(59, 12.7000)
(60, 13.1300)
(61, 12.7900)
(62, 12.7000)
(63, 13.1800)
(64, 13.0000)
(65, 12.6700)
(66, 13.0700)
(67, 13.2200)
(68, 13.2400)
(69, 12.8300)
(70, 13.1000)
(71, 12.9400)
(72, 12.8600)
(73, 13.0000)
(74, 12.6800)
(75, 12.5400)
(76, 12.6500)
(77, 13.1700)
(78, 13.2800)
(79, 13.0600)
(80, 12.7000)
(81, 12.6400)
(82, 13.0500)
(83, 12.8900)
(84, 13.1000)
(85, 12.7100)
(86, 12.6400)
(87, 12.4800)
(88, 12.7400)
(89, 12.6900)
(90, 12.6200)
(91, 12.7900)
(92, 12.8100)
(93, 13.1800)
(94, 12.2600)
(95, 11.9900)
(96, 12.0200)
(97, 12.5400)
(98, 12.4800)
(99, 12.2200)
};
    
\legend{R1, R2, R4, R16, R64, R128}
    
\end{axis}
\end{tikzpicture}
        }
        % \caption{}
        % \label{}
    \end{subfigure}
    \begin{subfigure}[b]{0.28\textwidth}
        \centering
        \resizebox{\textwidth}{!}{
            \input{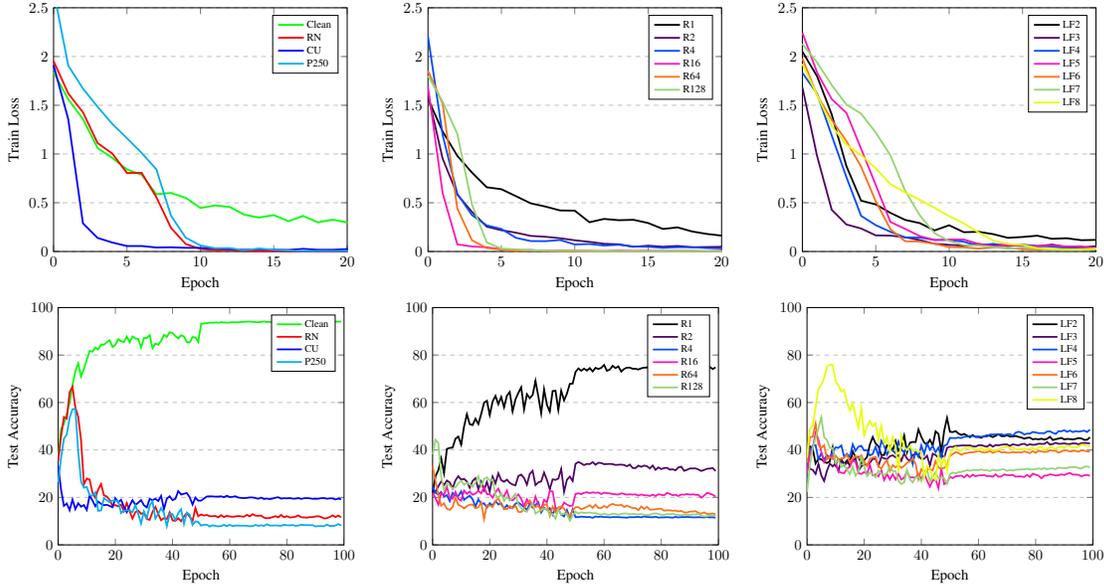}
        }
        % \caption{}
        % \label{}
    \end{subfigure}
    \caption{Train Loss and Test Accuracy curves when training a ResNet-18 on a variety of poisons. The first column displays curves for a clean training run plus three selected poisons: class-wise random noise (RN), class-wise unlearnable (CU), and 250-step PGD (P250). The second column displays curves for regions poisons. The third column displays curves for low-frequency poisons.}
    \label{fig:aggregate-traintest-curves}
    \vspace{-0.3cm}
\end{figure*}

\section{Experiments}

Employing a variety of poisons, our experiments were designed to answer two key questions: Are poisons that are learned faster a better defense against early stopping? and do stronger, more transferable adversarial attacks make more effective poisons?
In \cref{subsection:poisons-learned-faster}, we train a number of ResNet-18 models on different poisons with cross-entropy loss for $100$ epochs using a batch size of $128$. For our optimizer, we use SGD with momentum of $0.9$ and weight decay of \num{5e-4}. We use an initial learning rate of $0.1$, which decays by a factor of $10$ on epoch $50$.

\subsection{Poisons learned faster are more effective}
\label{subsection:poisons-learned-faster}

While poison effectiveness is generally measured by the final test set accuracy of a poisoned model, an arbitrarily low final test set accuracy is not effective if the victim could simply stop training early and obtain better test set performance. Thus, we define peak test accuracy as the highest test set performance over 100 epochs, and evaluate our poisons from \cref{tab:summary-all-poisons} on this metric. As we show in \cref{fig:scatter-peak-acc-vs-final-acc}, there are a number of poisons which reach low final test accuracy, but high peak test accuracy. 

% Note: 24 of 26 poisons reach loss 0.5 by epoch 10
To define the speed at which a poison is learned, we arbitrarily choose a loss threshold of $0.5$ and take note of the first epoch when the training loss is less than $0.5$. This is a reasonable loss threshold given that nearly $85\%$ of poisons we tested reach a cross-entropy loss of 0.5 by epoch 10. By taking note of the epoch after which a model reaches this threshold, we obtain a proxy for the ease with which a poison is learned.

For every poison, we hence obtain both an epoch until the loss threshold is reached and the overall peak test accuracy during training. We plot our results in \cref{fig:scatter-peak-acc-vs-epoch-threshold} and observe that poisons which are learned faster reach lower peak test accuracy, and are thus better defended against early-stopping. We provide training loss and validation accuracy curves in \cref{fig:aggregate-traintest-curves}. Our version of the sample-wise adversarial poison from Fowl \etal~\cite{fowl_adversarial_2021}, which we label \textbf{P250}, reaches a final test set accuracy of $8.15\%$, but reaches a peak test accuracy of $57.3\%$ as shown in \cref{fig:aggregate-traintest-curves} --- the poison is $7\times$ less effective when we stop training at epoch $8$! This serves as an example that poisons should be evaluated by their worst-case performance. The correlation between how quickly a poison is learned and its peak test accuracy suggests we should strive to minimize the loss early in training. 

\begin{table*}[ht]
  \centering
  \resizebox{0.70\textwidth}{!}{
  \begin{tabular}{l|ccccr|cc}
    \toprule
    \multirow{3}{*}{Poison} & \multicolumn{4}{c}{Victim Networks} &  & \multicolumn{2}{c}{Poison Detail} \\
        & ResNet-18 & VGG-19 & MobileNet & GoogLeNet & Avg ($\uparrow$) & Peak Acc ($\downarrow$) & Final Acc ($\downarrow$) \\
    \midrule
    RN   & 8.35 & 6.28  & 23.50 & 10.43 & 12.14 & 66.58 & 11.64 \\
    P10  & 14.24 & 7.55  & 34.55 & 13.00 & 17.34 & 83.43 & 61.84 \\
    P100 & 83.90 & 65.55 & 90.03 & 81.57 & 80.26 & \textbf{45.02} & 11.59 \\
    P250 & 95.55 & 84.58 & 95.73 & 90.93 & \textbf{91.70} & 57.30 & \textbf{8.15} \\
    M10  & 38.40 & 20.21 & 46.56 & 29.35 & 33.63 & 71.18 & 25.27 \\
    M100 & 53.18 & 22.79 & 46.75 & 43.12 & 41.46 & 69.05 & 14.58 \\
    M250 & 48.70 & 20.65 & 46.71 & 36.42 & 38.12 & 70.20 & 12.36 \\
    
    FC   & 94.95  & 38.88 & 53.88 & 50.85 & 59.64 & 87.09 & 86.93 \\
    FCS  & 99.97  & 73.18 & 76.23 & 88.69 & 84.52 & 83.82 & 83.37  \\
    FCSD & 100.00 & 83.53 & 86.24 & 94.74 & \textbf{91.13} & 80.31 & 80.00  \\
    \bottomrule
  \end{tabular}
  } % end resizebox
  \caption{Attack transferability of different error-maximizing CIFAR-10 poisons against victim models of varying architecture. Peak and Final accuracy ($\%$) are for a ResNet-18 trained on the poison. Values within $1\%$ of the best are in bold.}
  \label{tab:attack-transferability}
  \vspace{-0.3cm} % otherwise there is a weird gap below "4. Experiments" :(
\end{table*}

Our synthetic regions poisons and low-frequency poisons are a first step at designing perturbations which are learned quickly by convolutional DNNs like ResNet-18. Attempting to take advantage of a CNN's bias for spatially-local image features, our regions poisons have patches of the same color. Taking advantage of a DNNs bias for low-frequency patterns, our low frequency poisons span a spectrum of complexity. Training curves from \cref{fig:aggregate-traintest-curves} demonstrate that as the number of regions increases, poisons are generally learned more quickly, but there is a sweet spot for the number of regions (\textbf{R4} and \textbf{R16} have particularly low peak accuracy and final test accuracy). Compared to regions poisons, low frequency poisons perform more poorly overall. Interestingly, as the number of frequencies increases from \textbf{LF2} to \textbf{LF8}, the poisons appear to be learned more slowly, and the corresponding test accuracy curves show higher peak accuracies. For example, \textbf{LF7} reaches a peak accuracy of $53.72\%$ while \textbf{LF8} reaches $75.96\%$. Further work is needed to develop a poison which can induce lower than \textbf{R4}'s $24.51\%$ peak test accuracy on CIFAR-10.

\subsection{Stronger attacks do not make stronger poisons}
\label{subsection:stronger-attacks}

Both error-maximizing noise and error-minimizing noises were evaluated in Huang \etal.~\cite{huang_unlearnable_2021}, but their proposed error-minimizing sample-wise noise produced lower clean test set accuracy across the entire training process compared to other all other poisons. To optimize their error-maximizing noise, they used a 20-step PGD attack. 

Despite the apparent success of error-minimizing noise over error-maximizing noise, Fowl \etal. found a way to bring error-maximizing noise back into the spotlight: using a stronger 250-step PGD attack \cite{fowl_adversarial_2021}. Why did a 250-step attack behave so differently as a poison, when compared to the 20-step PGD attack? The findings of Fowl \etal~\cite{fowl_adversarial_2021} could be used as an indication that stronger adversarial attacks are the key to improving error-maximizing poisoning performance. Rather than continue to increase the number of attack steps, we opt to investigate whether adversarial attack transferability is correlated with poisoning performance. 

To evaluate this claim, we conduct an analysis of adversarial attack transferability across a range of error-maximizing attacks from \cref{tab:summary-all-poisons}. We consider four DNN architectures as victim networks: a ResNet-18 \cite{he2016deep}, VGG-19 \cite{Simonyan2015VeryDC}, MobileNet \cite{Howard2017MobileNetsEC}, and GoogLeNet \cite{Szegedy2015GoingDW}, which we train on CIFAR-10 for 200 epochs. Recall that sample-wise error-maximizing poisons are constructed by performing an adversarial attack on each clean image from the training set. Thus, attack transferability is measured by computing the fraction of examples in the poison which are misclassified by the victim network. We present attack transferability results alongside the corresponding poison's peak and final test accuracy in \cref{tab:attack-transferability}. 

An adversarial attack can be considered stronger than another if it uses more attack steps or if it is more transferable to other DNNs of differing architectures. In terms of peak test accuracy, increasing the number of attack steps for a PGD or MI-FGSM attack does not improve peak test accuracy. For example, \textbf{P100} achieves a peak test accuracy of $45.02\%$, while the more transferable \textbf{P250} poison achieves $57.30\%$. Stronger, more transferable adversarial attacks do not necessarily degrade test accuracy either. The case of \textbf{FCSD} is a striking example: the adversarial attack is one of the most transferable across the four victim networks, yet performs very poorly as a poison.

\section{Conclusion}

In this paper, we made a number of observations which should be taken into account when designing poisons for the purpose of data privacy protection. Motivated by defending against the mitigating effect of early-stopping, we found that poison training exhibits a correlation between how quickly a model reaches a low loss threshold and the peak test accuracy overall. This leads us to propose that poisons should be learned quickly to ensure peak test accuracy remains low throughout training and user privacy is meaningfully protected.

To further understand this phenomenon, we craft two kinds of synthetic poisons aimed at taking advantage of CNN biases. But while random, spatially-local regions poisons are learned quickly, they suffer from being slightly perceptible. Low frequency perturbations, on the other hand, do not achieve low test accuracy.

Finally, our experiments suggest that adversarial attacks which are more transferable or stronger in terms of steps do not lead to effective error-maximizing poisons. Due to high peak test accuracy, we caution against error-maximizing noises for data poisoning.

% Finally, we discuss potential properties that make data unlearnable by DNN models.

%To further demonstrate the generality of our findings, we repeat our poison training experiments on architectures like MLP, ConvMixer, and ViT. 
\textbf{Acknowledgements.} This work is supported in part by the Guaranteeing AI Robustness Against Deception (GARD) program from DARPA. Pedro is supported by an Amazon Lab126 Diversity in Robotics and AI Fellowship.

%%%%%%%%% REFERENCES
{\small
\bibliographystyle{ieee_fullname}
\bibliography{mylibrary}
}

\end{document}